\documentclass{article} 
\usepackage{iclr2026_conference,times}

\usepackage{amsmath,amsfonts,bm}

\def\eqref#1{equation~\ref{#1}}

\def\1{\bm{1}}

\DeclareMathAlphabet{\mathsfit}{\encodingdefault}{\sfdefault}{m}{sl}
\SetMathAlphabet{\mathsfit}{bold}{\encodingdefault}{\sfdefault}{bx}{n}

\usepackage[utf8]{inputenc} 
\usepackage[T1]{fontenc}    
\usepackage[pagebackref=true,breaklinks,colorlinks,citecolor=blue,linkcolor=blue,urlcolor=blue,hypertexnames=false]{hyperref}
\usepackage{url}            
\usepackage{booktabs}       
\usepackage{amsfonts}       
\usepackage{nicefrac}       
\usepackage{microtype}      
\usepackage{xcolor}         
\usepackage{tabularx}
\usepackage{subcaption}
\usepackage{multicol}
\usepackage{multirow}
\usepackage{tabularray}
\usepackage{amsmath}
\usepackage{algorithm}%
\usepackage{algorithmicx}%
\usepackage{algpseudocode}
\usepackage{amsfonts}
\usepackage{siunitx}
\usepackage{colortbl}
\usepackage{graphicx}
\usepackage{lipsum}
\usepackage{nicefrac}
\usepackage{tabularray}
\usepackage{wrapfig}
\usepackage{pifont}
\usepackage{xspace}         

\newcommand{\ourmethod}{MIRAGE }
\usepackage{pgfplots,gincltex}
\usepgfplotslibrary{groupplots}
\pgfplotsset{compat=1.6}
\usepackage{tikz}
\usetikzlibrary{arrows.meta}

\definecolor{pltblue}{RGB}{174, 199, 232}
\definecolor{pltorange}{RGB}{255, 229, 204}
\definecolor{pltgreen}{RGB}{204, 229, 204}
\definecolor{pltred}{RGB}{229, 204, 204}
\definecolor{pltpurple}{RGB}{239, 218, 230}

\definecolor{tabblue}{HTML}{1f77b4}
\definecolor{taborange}{HTML}{ff7f0e}
\definecolor{tabgreen}{HTML}{2ca02c}
\definecolor{tabred}{HTML}{d62728}
\definecolor{tabpurple}{HTML}{9467bd}

\definecolor{cblue}{RGB}{173, 201, 233}
\definecolor{clblue}{RGB}{222, 234, 246}
\definecolor{corange}{RGB}{255, 152, 67}
\definecolor{lorgange}{RGB}{255, 221, 149}

\newcommand{\cc}[1]{\cellcolor{clblue!50}{#1}}

\newcommand\sotaa{\textcolor{tabred}}
\newcommand\sotab{\textcolor{tabblue}}

\def\eg{\emph{e.g.}\xspace} 
\def\ie{\emph{i.e.}\xspace}

\def\etal{\emph{et al.}\xspace}
\def\sota{state-of-the-art\xspace}

\usepackage{listings}
\definecolor{codeblue}{rgb}{0.25,0.5,0.5}
\definecolor{codekw}{rgb}{0.85, 0.18, 0.50}
\title{Efficient Degradation-agnostic Image Restoration via Channel-Wise Functional Decomposition and Manifold Regularization}

\author{  
Bin Ren\textsuperscript{1,2}\quad 
Yawei Li\textsuperscript{3}\quad  
Xu Zheng\textsuperscript{4}\quad 
Yuqian Fu\textsuperscript{5}\quad 
Danda Pani Paudel\textsuperscript{5}\quad
Hong Liu\textsuperscript{6$\star$}\\ 
\textbf{~Ming-Hsuan Yang}\textsuperscript{7}\quad 
\textbf{Luc Van Gool}\textsuperscript{5}\quad 
\textbf{Nicu Sebe}\textsuperscript{2}\\
\normalsize{
\textsuperscript{1}Mohamed bin Zayed University of Artificial Intelligence\quad 
\textsuperscript{2}University of Trento}\quad
\textsuperscript{3}ETH Z\"urich\\
\normalsize{
\textsuperscript{4}HKUST (GZ)\quad
\textsuperscript{5}INSAIT, Sofia University ``St. Kliment Ohridski''}\\
\normalsize{~\textsuperscript{6}Peking University\quad 
\textsuperscript{7}
University of California, Merced} \\
}

\iclrfinalcopy 
\begin{document}

{
    \renewcommand{\thefootnote}{\fnsymbol{footnote}}
    \footnotetext[1]{~indicates corresponding author: Hong Liu \textless\href{hongliu@pku.edu.cn}{hongliu@pku.edu.cn}\textgreater.%
}

\maketitle
\begin{center}
    \vspace{-5mm}
    \includegraphics[width=0.99\linewidth]{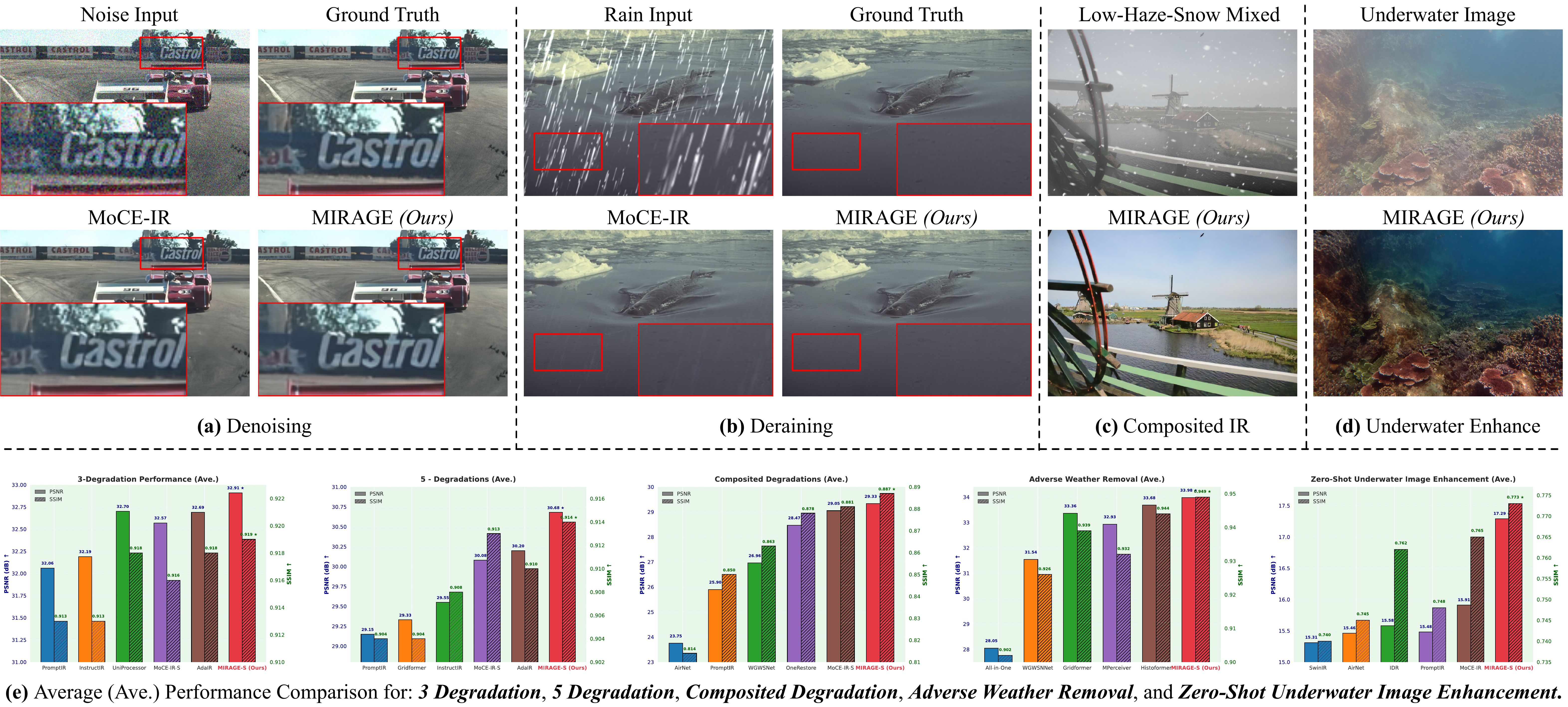}%
    \vspace{0mm}
    \captionof{figure}{
    \textit{(a)-(d)}: Visual comparison for Denoising, Deraining, Composited Degradations (low-light, haze, and snow), and underwater image enhancement.
    \textit{(e)}: The average PSNR and SSIM comparison across 4 challenging all-in-one and 1 zero-shot settings (Please zoom in for a better view).
    }
    \label{fig:teaser}
\end{center}%

\begin{abstract}
    Degradation-agnostic image restoration aims to handle diverse corruptions with one unified model, but faces fundamental challenges in balancing efficiency and performance across different degradation types. Existing approaches either sacrifice efficiency for versatility or fail to capture the distinct representational requirements of various degradations. We present MIRAGE, an efficient framework that addresses these challenges through two key innovations. First, we propose a channel-wise functional decomposition that systematically repurposes channel redundancy in attention mechanisms by assigning CNN, attention, and MLP branches to handle local textures, global context, and channel statistics, respectively. This principled decomposition enables degradation-agnostic learning while achieving superior efficiency-performance trade-offs. Second, we introduce manifold regularization that performs cross-layer contrastive alignment in Symmetric Positive Definite (SPD) space, which empirically improves feature consistency and generalization across degradation types. Extensive experiments demonstrate that MIRAGE achieves state-of-the-art performance with remarkable efficiency, outperforming existing methods in various all-in-one IR settings while offering a scalable and generalizable solution for challenging unseen IR scenarios.
\end{abstract}
\section{Introduction}
\label{sec:introduction}
Image Restoration (IR) aims to recover clean images from inputs degraded by diverse real-world corruptions such as noise, blur, haze, rain, and low-light conditions~\citep{zamir2022restormer,li2023efficient,ren2024sharing,potlapalli2023promptir}. A central challenge is \textit{degradation-agnostic restoration}: developing a single model that can generalize across heterogeneous degradations. Despite recent progress, existing approaches often face an efficiency–performance dilemma. On the one hand, heavyweight designs based on prompts, instructions, or large vision–language models provide versatility but incur substantial computational cost~\citep{potlapalli2023promptir,zamfir2025moce,jiang2024survey}. On the other hand, lightweight solutions improve efficiency at the expense of restoration quality~\citep{li2022all,tang2025degradation}. Achieving both robustness and efficiency within a unified framework remains an open problem.

This difficulty can be better understood from two complementary perspectives. First, different degradation types impose fundamentally different representational requirements: additive corruptions (\eg, noise, rain) primarily affect local textures, multiplicative distortions (\eg, haze, low-light) require global context modeling, and kernel-based degradations (\eg, blur) call for multi-scale structural reasoning. At the same time, basic architectural modules exhibit distinct inductive biases: convolutional filters excel at local texture modeling, attention mechanisms capture long-range dependencies, and MLPs enhance channel statistics. This motivates the insight that \textit{an effective restoration model should systematically align distinct modules with complementary representational functions}. Second, recent studies reveal substantial redundancy in attention-based models, particularly along the channel dimension~\citep{venkataramanan2023skip,dong2021attention}. Many channels encode overlapping information, suggesting that this redundancy could be \textit{repurposed} rather than discarded. Leveraging this observation allows for architectures that remain compact while preserving expressive capacity.
These observations highlight that unified IR benefits not only from adding new modules, but from a principled reorganization of existing capacity based on redundancy patterns and complementary inductive biases. 
This perspective motivates our design philosophy in MIRAGE, where representational roles are explicitly aligned with structural evidence rather than heuristic module stacking.

Building on these insights, we present MIRAGE, an efficient framework for degradation-agnostic image restoration. MIRAGE introduces two components. (i) \textit{Channel-wise functional decomposition}, where the input feature map is partitioned along the channel dimension and processed by three specialized branches: convolution for local textures, attention for global context, and MLP for channel statistics. This structured decomposition repurposes redundant capacity into complementary roles, yielding both interpretability and strong efficiency–performance trade-offs. (ii) \textit{Manifold regularization}, a cross-layer contrastive strategy that leverages natural feature pairs within the model. Inspired by deeply supervised networks~\citep{lee2015deeply}, we hypothesize that natural contrastive pairs exist between shallow and latent representations. Shallow features preserve fine spatial details but are sensitive to noise, while latent features are more abstract and semantically stable; aligning them encourages more robust shared representations. Importantly, rather than computing contrastive loss in Euclidean space, which may distort similarity when comparing structured representations, we operate in the Symmetric Positive Definite (SPD) manifold space. This formulation provides a more faithful alignment of representations, leading to improved generalization across degradation types.  
Overall, MIRAGE provides a structurally grounded view of unified IR, where representational capacity is allocated and aligned based on statistical evidence at both the spatial and depth levels.

Extensive experiments across five degradation settings show that MIRAGE achieves state-of-the-art performance with remarkable efficiency: our model has only 6M parameters, more than five times smaller than recent prompt-based baselines, while also generalizing well to unseen scenarios such as underwater image enhancement. Both the visual and per-setting PSNR results are shown in Fig.~\ref{fig:teaser}.  

Our contributions are summarized as follows:
\begin{itemize}
    \item We propose a principled channel-wise functional decomposition strategy that aligns convolution, attention, and MLP with distinct representational roles, enabling efficient and effective degradation-agnostic restoration.  
    \item We introduce manifold regularization through cross-layer contrastive alignment between shallow and latent features. We exploit natural contrastive pairs within the model, and perform this alignment in the SPD manifold space rather than Euclidean space, providing more faithful representation similarity and improved generalization across diverse degradations.  
    \item We conduct comprehensive experiments across single, mixed, and unseen degradation settings, establishing MIRAGE as a strong and practical baseline for all-in-one IR.  
\end{itemize}
\section{Related Work}
\label{sec:related-work}
\textbf{Image Restoration with Various Architectures.} 
IR addresses the ill-posed problem of retoring high-quality images from degraded inputs and has long been a core task in computer vision with broad applications~\citep{richardson1972bayesian,banham1997digital,xie2025mat,li2023lsdir,zamfir2024details}. Early methods relied on model-based formulations with handcrafted priors, but deep learning has shifted the field toward data-driven approaches, including regression-based~\citep{lim2017enhanced,lai2017deep,liang2021swinir,chen2021learning,li2023efficient,zhang2024transcending} and generative pipelines~\citep{gao2023implicit,wang2022zero,luo2023image,yue2023resshift,zhao2024denoising}. These methods employ diverse backbones: convolutional networks for local structures~\citep{dong2015compression,zhang2017learning,zhang2017beyond,wang2018recovering}, MLPs and state space models for channel or sequential dependencies~\citep{tu2022maxim,guo2024mambair,zhu2024vision,mamba,mamba2,tang2025ramir}, and Transformers for long-range interactions~\citep{liang2021swinir,ren2023masked,li2023efficient,zamir2022restormer,dosovitskiy2020image,liu2023spatio,shi2025vmambair}, achieving promising results. Despite these advances, most IR solutions remain degradation-specific, addressing tasks such as denoising~\citep{zhang2019residual}, dehazing~\citep{wu2021contrastive}, deraining~\citep{jiang2020multi}, or deblurring~\citep{kong2023efficient}, motivating the need for unified frameworks that generalize across diverse degradations while remaining efficient.

\textbf{Degradation-agnostic Image Restoration.} 
While training task-specific models for individual degradations can be effective, it is impractical to maintain separate models for each corruption. Real-world images often suffer from mixed degradations, making independent treatment infeasible, and task-specific approaches further increase computational and storage costs, amplifying their environmental footprint.
To overcome these limitations, the emerging field of degradation-agnostic IR focuses on single-blind models capable of handling multiple degradation types simultaneously~\citep{zamfir2025moce,zeng2025vision,zheng2024learning,ren2026any}.
For example, AirNet~\citep{li2022all} achieves blind All-in-One image restoration by using contrastive learning to derive degradation representations from corrupted images, which are then leveraged to reconstruct clean images. Building on this, IDR~\citep{zhang2023ingredient} tackles the problem by decomposing degradations into fundamental physical components and applying a two-stage meta-learning strategy.
More recently, the extra learnable prompt-based paradigm~\citep{potlapalli2023promptir,wang2023promptrestorer,li2023prompt,tian2025degradation} has introduced a visual prompt learning module, enabling a single model to better handle diverse degradation types by leveraging the discriminative capacity of learned visual prompts. Extending this idea, some works further model prompts from a frequency perspective~\citep{cui2025adair} or propose more complex architectures with additional datasets~\citep{dudhane2024dynamic}.
However, visual prompt modules often result in increased training time and decreased efficiency~\citep{cui2025adair}. 
Meanwhile, inspired by recent advances in self-supervised learning, several works~\citep{wu2021contrastive,chen2022unpaired} have explored contrastive objectives to enhance low-level representations, though mainly within single-task IR scenarios. 
For the degradation-agnostic setting~\citep{jiang2024survey,li2022all,chen2025multi,zhang2025perceive}, the most recent DA-RCOT~\citep{tang2025degradation} introduces a contrastive loss applied to residual feature space, illustrating that contrastive signals can also benefit unified IR models. 
In contrast, our work aims to improve the model’s ability to capture representative degradation cues within the SPD space without relying on heavy or complex prompt designs. Our goal in this work is to develop a degradation-agnostic image restorer that remains both computationally efficient and environmentally sustainable.
\section{Preliminary: Degradation-Aware Architectures for IR}
\label{sec:preliminary}
\textbf{Image Degradation and Restoration.}
Image restoration seeks to recover a clean image $\mathbf{x}$ from a degraded observation $\mathbf{y}$:
\begin{equation}
\mathbf{y} = \mathcal{D}(\mathbf{x}) + \mathbf{n},
\end{equation}
where $\mathcal{D}(\cdot)$ denotes a degradation operator and $\mathbf{n}$ noise. Real-world degradations are diverse—additive (\eg, Gaussian noise, rain: $\mathbf{y} = \mathbf{x} + \mathbf{n}$), multiplicative (\eg, haze, speckle: $\mathbf{y} = \mathbf{x} \cdot \mathbf{m}$), or convolutional (\eg, blur, super-resolution: $\mathbf{y} = \mathbf{k} * \mathbf{x} + \mathbf{n}$)~\citep{he2025diffusion}. These factors often co-occur and are spatially variant~\citep{zhai2023comprehensive}, forming compound pipelines:
\begin{equation}
\mathbf{y} = \mathcal{D}_3\big(\mathcal{D}_2(\mathcal{D}_1(\mathbf{x}))\big) + \mathbf{n}.
\end{equation}
Such complexity demands models that preserve local details while reasoning about global structures.

\textbf{Architectural Biases for Degradation Modeling.}
Deep networks embody distinct inductive biases:
\emph{CNNs} capture local spatial patterns:
$\mathbf{y}p = \sum{i \in \mathcal{N}(p)} w_i \cdot \mathbf{x}_i$,
effective for uniform or spatially invariant degradations.
\emph{Transformers} exploit global self-attention:
$\mathbf{y}i = \sum_j \alpha{ij} \cdot \mathbf{V}_j$,
well-suited for non-uniform, structured degradations (\eg haze, patterned noise).
\emph{MLPs}, especially token-mixing forms, apply flexible position-wise mappings:
$\mathbf{y} = \mathbf{W}_2 \cdot \phi(\mathbf{W}_1 \cdot \mathbf{x})$,
though with weak spatial priors.

Each paradigm shows strengths yet clear limitations—CNNs excel in local fidelity, Transformers in global reasoning, and MLPs in flexible feature interactions, but lack inductive structure. Alone, they are insufficient for complex degradations and often parameter-heavy. Their complementarity motivates unified, degradation-aware architectures that leverage all three for robust IR in the wild.
\section{The Proposed \ourmethod}
\label{sec:method}
\begin{figure}[!t]
    \centering
    \includegraphics[width=0.97\linewidth]{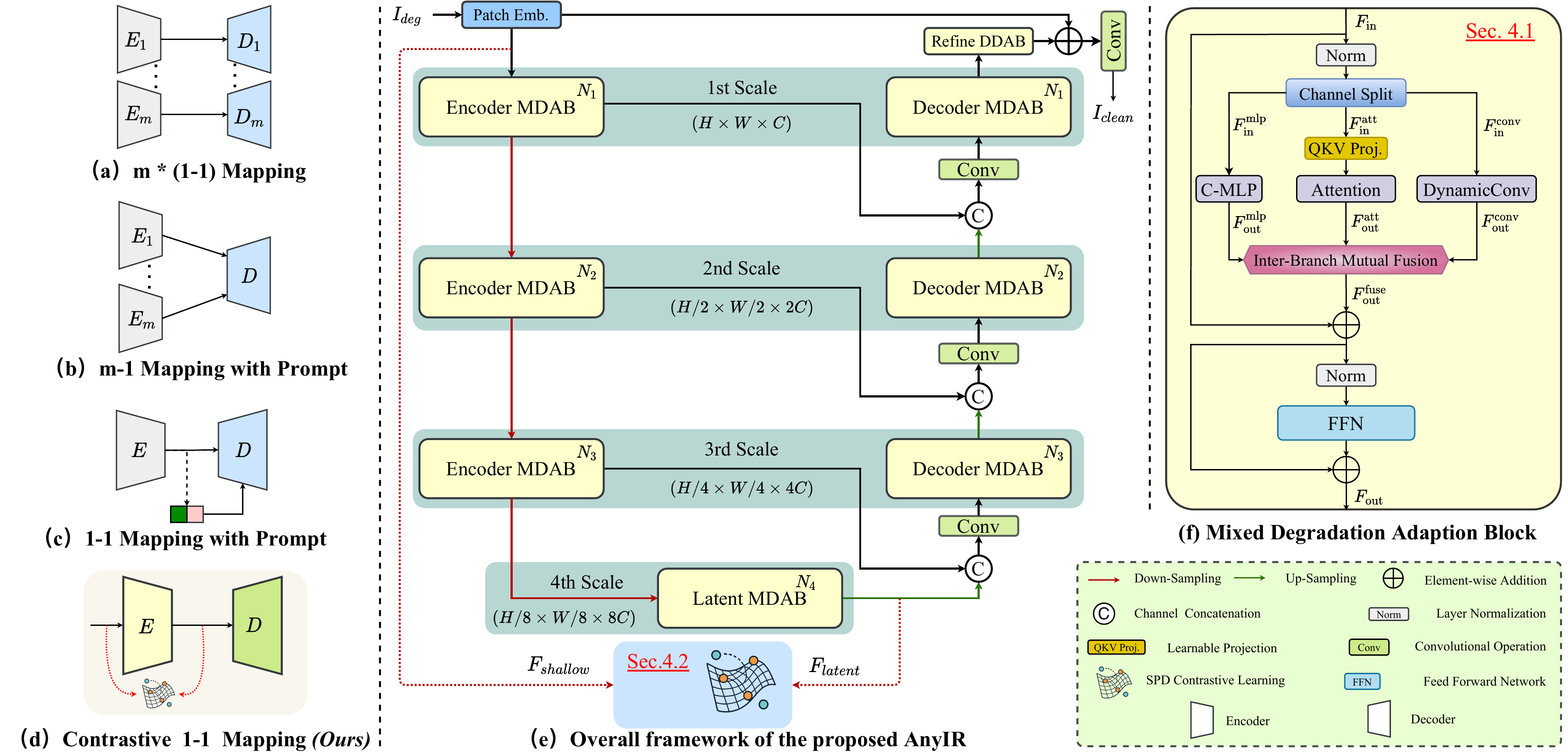}
    \vspace{-1mm}
    \caption{
    \textit{(a)-(c)}: The most adopted all-in-one image restoration encoder-decoder pipelines.
    \textit{(d)}: The toy illustration of our SPD contrastive pipeline. 
    \textit{(e)}: The overall framework of the proposed {\ourmethod}: \ie, a convolutional patch embedding, a U-shape encoder-decoder main body, an extra refined block, and the proposed SPD contrastive learning algorithm. 
    \textit{(f)}: Structure of each mixed degradation adaptation block (MDAB).}
    \label{fig:framework}
    \vspace{-1mm}
\end{figure}

The design of MIRAGE is guided by two empirical observations. 
\textit{(i)} Attention features consistently exhibit low-rank channel redundancy across scales (Fig.~\ref{fig:redundency}), indicating that a non-trivial portion of the representational capacity can be reassigned without loss of expressiveness. 
\textit{(ii)} Different degradations favor complementary inductive biases, \ie, local texture sensitivity, global contextual aggregation, and channel-statistical modulation. 
These observations motivate a principled partition of feature channels into convolutional, attention, and MLP pathways, allowing each subspace to specialize in the bias it is best suited for while maintaining overall model compactness. 
In parallel, the depth-asymmetric covariance structures of shallow and latent representations provide a natural basis for cross-layer alignment, for which the SPD formulation offers a geometry-preserving representation.

Prior works either train a separate model per degradation (Fig.~\ref{fig:framework}a), adopt multi-encoder–single-decoder designs that inflate parameters (Fig.~\ref{fig:framework}b), or rely on large-scale prompt-based models with visual/textual cues (Fig.~\ref{fig:framework}c).
In contrast, we propose a simple yet effective mixed-backbone architecture (Fig.~\ref{fig:framework}d), which already forms a strong restoration baseline (Sec.~\ref{subsec:mix_backbone}) and is further enhanced by cross-layer contrastive learning in SPD space between shallow and latent features (Sec.~\ref{subsec:spd_ctrs}).

\subsection{Mixed Degradation Adaptation Block for Degradation-Agnostic IR}
\label{subsec:mix_backbone}
\begin{wrapfigure}[16]{r}{8cm}
    \centering
    \vspace{0mm}
    \includegraphics[width=0.98\linewidth]{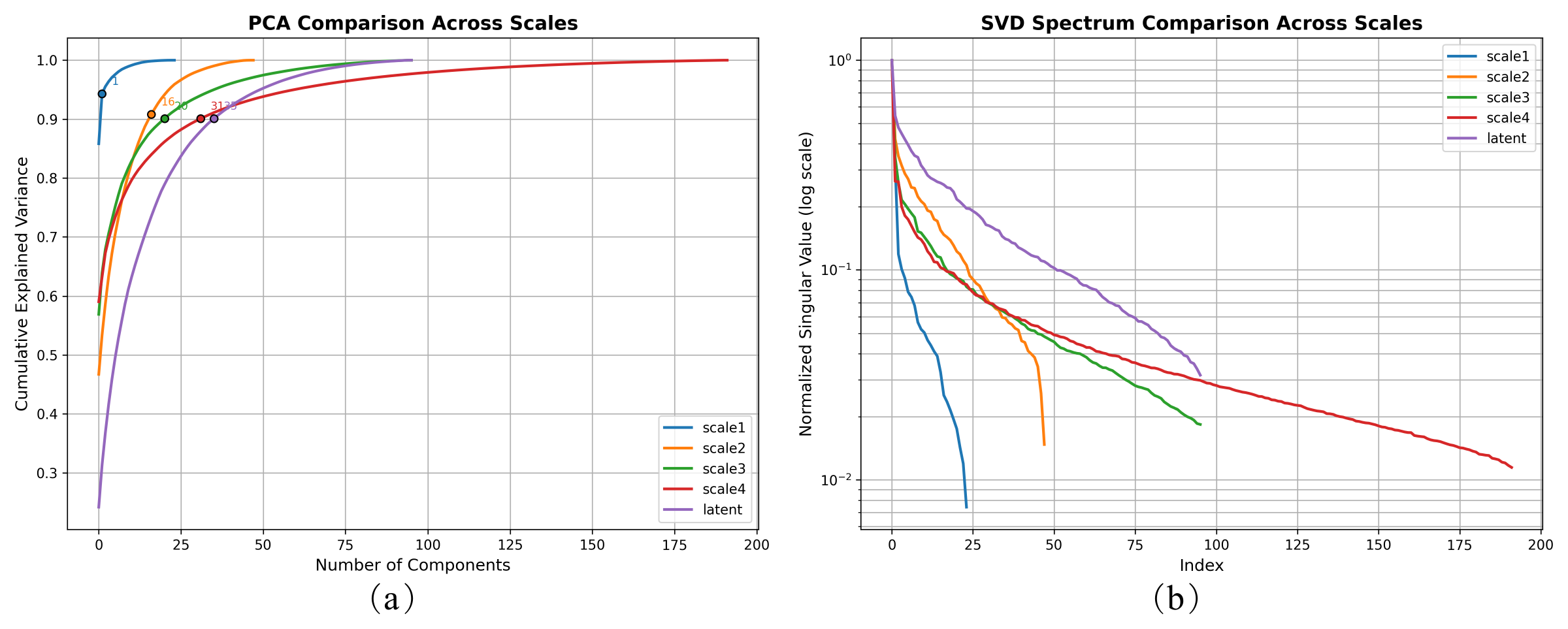}
    \vspace{-3mm}
    \caption{
    Channel redundancy analysis across multiple feature scales.
    (a) Cumulative explained variance curves from PCA applied to the channel dimension of features from 1-4 scales and one latent scale.
    (b) Normalized singular value spectra (in log scale) of the same features via SVD. Latent feature in both plots means the channel-wise projected 4th Scale feature.
    }
    \label{fig:redundency}
    \vspace{-3mm}
\end{wrapfigure}
\textbf{Redundancy in MHAs Opens Opportunities for Hybrid Architectures.}
Redundancy has long been recognized as a fundamental limitation in multi-head self-attention (MHA), the core building block of Transformers, in both NLP and vision domains~\citep{nguyen2022improving,nguyen2022improvingkeys,xiao2024improving,brodermann2025cafuser,wang2022improved,venkataramanan2023skip}.
Prior studies indicated that not all attention heads contribute equally, \ie, some are specialized and crucial, while others can be pruned with negligible impact.
\textit{This inherently implies redundancy in the channel dimension, as MHA outputs are concatenated along this axis.}
To empirically verify this redundancy in the context of IR, we analyze intermediate features from a lightweight attention-only model (details in the Appendix~\ref{fig:supp:ab_mdab}). 
Specifically, we compute cumulative explained variance via PCA and normalized singular value spectra via SVD across multiple feature scales.
Fig.~\ref{fig:redundency}(a) shows earlier scales (\eg, 1st Scale) need far fewer principal components to retain most variance, suggesting high redundancy. Fig.~\ref{fig:redundency}(b) further supports this, with a sharper singular value decay at shallower stages, indicating stronger low-rank structure in channel-wise representations. Even at the deepest stage (\eg, 4th Scale), achieving 90\% variance requires only 31 of 192 components ($\approx 16\%$), confirming redundancy persists throughout.

This insight motivates a departure from traditional head/channel pruning. Instead of discarding redundant capacity, we propose to \textit{repurpose} it by splitting the channel dimension into three parts and feeding them into distinct architectural branches, \ie, attention, convolution, and MLP. This hybrid formulation leverages complementary inductive biases and makes full use of available representational space, offering a principled and efficient alternative to the previous pure MSA-based designs.

\begin{algorithm}[t]
\caption{Mixed Parallel Degradation Adaptation}
\label{alg:mpda}
\begin{algorithmic}[1]
\Require $F_{\text{in}}^{\text{att}},\ F_{\text{in}}^{\text{conv}},\ F_{\text{in}}^{\text{mlp}}$ \Comment{Input features from three branches}
\Ensure $F_{\text{out}}$ \Comment{Final fused output}

\Statex \textbf{[Att] Attention Path}
\State $Q,K,V \gets \text{Linear}(F_{\text{in}}^{\text{att}})$ \Comment{Projection to attention tokens}
\State $F_{\text{out}}^{\text{att}} \gets \texttt{Softmax}(\frac{QK^\top}{\sqrt{d}})V$ \Comment{Multi-head self-attention}

\Statex \textbf{[Conv] Dynamic Convolution Path}
\State $F' \gets \text{Conv1x1}(\text{Norm}(F_{\text{in}}^{\text{conv}}))$ \Comment{Normalization and expansion}
\State $\gamma,\beta,\alpha \gets \text{Split}(F')$ \Comment{Gating, intermediate, convolutional paths}
\State $\alpha' \gets \text{DynamicDepthwiseConv}(\alpha)$ \Comment{Content-adaptive depthwise conv}
\State $\hat{F} \gets \sigma(\gamma/\tau) \cdot \text{Concat}(\beta, \alpha')$ \Comment{Gated local enhancement}
\State $F_{\text{out}}^{\text{conv}} \gets \text{Conv1x1}(\hat{F}) + F_{\text{in}}^{\text{conv}}$ \Comment{Residual projection}

\Statex \textbf{[MLP] MLP Path}
\State $F_{\text{out}}^{\text{mlp}} \gets \text{MLP}(F_{\text{in}}^{\text{mlp}})$ \Comment{Channel-wise transformation brings more non-linearity}

\Statex \textbf{[Fusion] Inter-Branch Mutual Fusion}
\State $F_{\text{out}}^{\text{att}'} \gets F_{\text{out}}^{\text{att}} + \lambda_{\text{att}} \cdot \sigma(F_{\text{out}}^{\text{conv}} + F_{\text{out}}^{\text{mlp}})$ \Comment{Fuse conv and MLP into attention}
\State $F_{\text{out}}^{\text{conv}'} \gets F_{\text{out}}^{\text{conv}} + \lambda_{\text{conv}} \cdot \sigma(F_{\text{out}}^{\text{att}} + F_{\text{out}}^{\text{mlp}})$ \Comment{Fuse attention and MLP into conv}
\State $F_{\text{out}}^{\text{mlp}'} \gets F_{\text{out}}^{\text{mlp}} + \lambda_{\text{mlp}} \cdot \sigma(F_{\text{out}}^{\text{att}} + F_{\text{out}}^{\text{conv}})$ \Comment{Fuse attention and conv into MLP}

\Statex \textbf{Output Projection}
\State $F_{\text{out}}^{\text{fuse}} \gets \text{Project}(\text{Concat}(F_{\text{out}}^{\text{att}'},\ F_{\text{out}}^{\text{conv}'},\ F_{\text{out}}^{\text{mlp}'}))$ \Comment{Final unified representation}

\State \Return $F_{\text{out}}^{\text{fuse}}$
\end{algorithmic}
\end{algorithm}

\textbf{Parallel Design Brings More Efficiency.}  
As shown in Lines 1–8 of Alg.~\ref{alg:mpda}, we instantiate this idea through a structurally parallel design that simultaneously exploits complementary inductive biases. As illustrated in Fig.~\ref{fig:framework}(f), the input feature \( F_{\text{in}} \in \mathbb{R}^{h \times w \times c} \) is evenly partitioned along the channel dimension into three sub-tensors (\ie, \( F_{\text{in}}^{\text{att}} \), \( F_{\text{in}}^{\text{conv}} \), and \( F_{\text{in}}^{\text{mlp}} \),), which are then processed in parallel by attention, convolution, and MLP branches. 
Each branch operates only on its allocated fraction of channels, substantially reducing computational cost, while its architectural heterogeneity enriches the representational space. This parallel decomposition achieves a favorable balance between efficiency and expressiveness, in contrast to prior designs that rely on purely attention-based processing.

\textbf{Inter-Branch Mutual Fusion Injects Expressivity Before FFN.} 
While the parallel design improves efficiency and modularity, it reduces interaction across branches. 
To mitigate this, Lines 9–13 of Alg.~\ref{alg:mpda} introduce an inter-branch fusion mechanism, where each branch is enhanced via gated aggregation of the rest, modulated by learnable coefficients \( \lambda \).  
This introduces cross-path context blending, reinforcing feature complementarity before unification, forming an effective pre-FFN encoder.

Compared to the attention-only models,
the fused output in Alg.~\ref{alg:mpda} introduces richer interactions. This enhances the model’s ability to fit complex degradation mappings, making it more suitable for mixed or ambiguous degradations.  
Subsequently, layer normalization, a feed-forward network (FFN), and a residual connection are applied:  
\( F_{\text{out}} = \operatorname{FFN}(\operatorname{Norm}(F_{\text{out}}^{\text{fuse}})) + F_{\text{out}}^{\text{fuse}} \).  
This sequence stabilizes feature distributions and further boosts expressiveness.

\subsection{Shallow-Latent Contrastive Learning via SPD Manifold Alignment}
\label{subsec:spd_ctrs}
The unified IR model requires a single backbone to process degradations that depend on fundamentally different representational levels. 
Shallow layers primarily encode degradation-specific, fine-grained structures, whereas deeper layers become more semantic and statistically stable. 
This inherent depth asymmetry introduces representation drift when multiple degradations share the same feature space, motivating a mechanism that explicitly enforces cross-stage consistency. 
We therefore treat shallow and latent features as complementary views of the underlying signal and align them to stabilize the shared representation space, thereby improving generalization across heterogeneous degradations.

\textbf{Shallow-Latent Feature Pairs are Naturally Contrastive Pairs.}  
Features extracted at different depths exhibit fundamentally different statistical properties. As shown in Fig.~\ref{fig:cross_layer_sim}, shallow-stage features (\eg, Scale1) present \textit{sparse and decorrelated channel distributions}, while deeper layers (\eg, Scale4) become \textit{increasingly redundant and concentrated}.   
This trend is quantitatively supported by the effective rank ratio across scales, which increases from only 4.2\% ($\nicefrac{1}{24}$ at 1st Scale) to 16.1\% ($\nicefrac{31}{192}$ at 4th Scale). 
However, by compressing the deep features through a lightweight MLP, we obtain a latent representation with a notably higher rank ratio of 36.5\% ($\nicefrac{35}{96}$), indicating a more decorrelated and expressive embedding.  
This structural disparity between sparse, localized shallow features and compressed, semantic latent ones naturally defines a contrastive pairing without requiring augmentation. We leverage this depth-asymmetric contrast to impose consistency across stages, enabling better semantic alignment and stronger representational generalization under complex degradation conditions. \textit{Note that this study is conducted under noise degradation; however, similar trends are consistently observed for other degradations as well}. See the appendix for more details.

\begin{figure}[!t]
    \centering
    \begin{subfigure}[t]{0.19\linewidth}
        \centering
        \includegraphics[width=\linewidth]{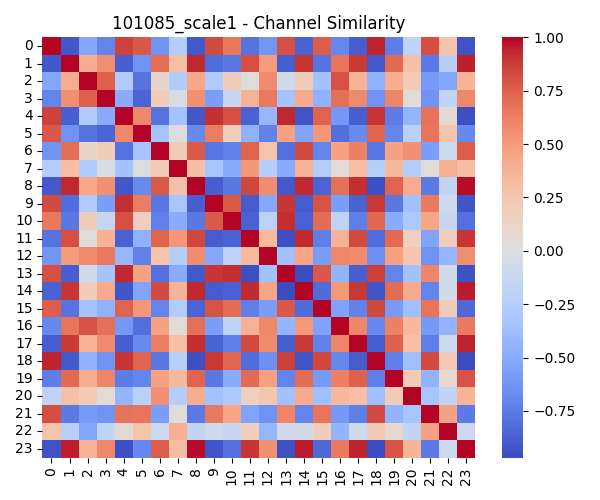}
        \vspace{-5mm}
        \caption{}
    \end{subfigure}
    \hfill
    \begin{subfigure}[t]{0.19\linewidth}
        \centering
        \includegraphics[width=\linewidth]{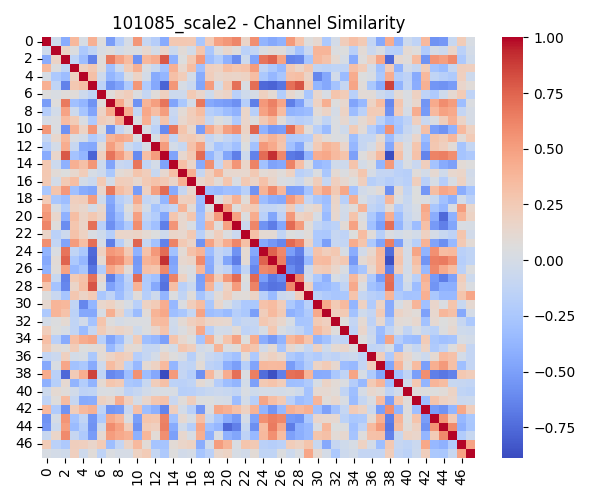}
        \vspace{-5mm}
        \caption{}
    \end{subfigure}
    \hfill
    \begin{subfigure}[t]{0.19\linewidth}
        \centering
        \includegraphics[width=\linewidth]{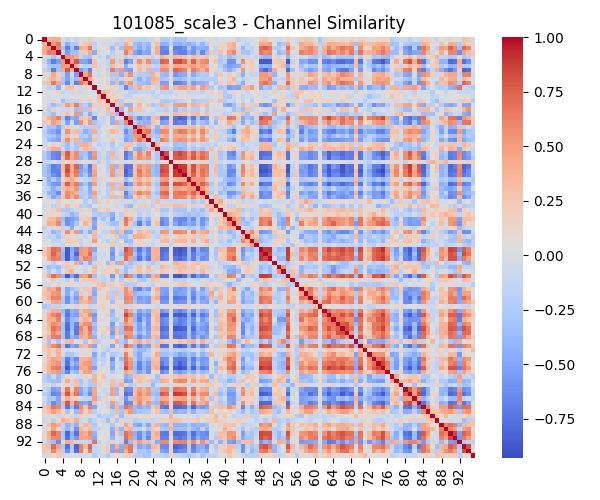}
        \vspace{-5mm}
        \caption{}
    \end{subfigure}
    \hfill
    \begin{subfigure}[t]{0.19\linewidth}
        \centering
        \includegraphics[width=\linewidth]{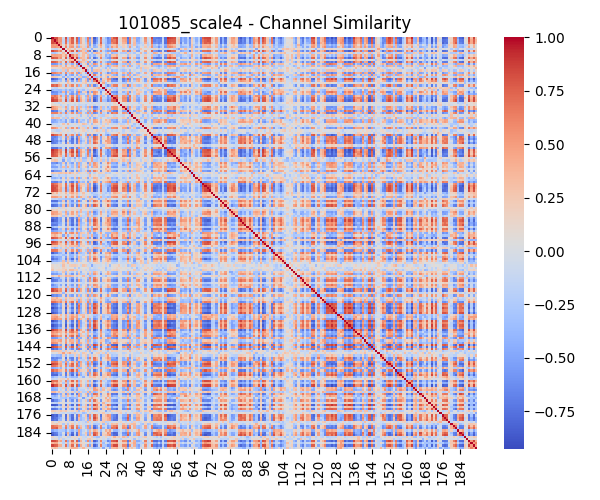}
        \vspace{-5mm}
        \caption{}
    \end{subfigure}
    \hfill
    \begin{subfigure}[t]{0.19\linewidth}
        \centering
        \includegraphics[width=\linewidth]{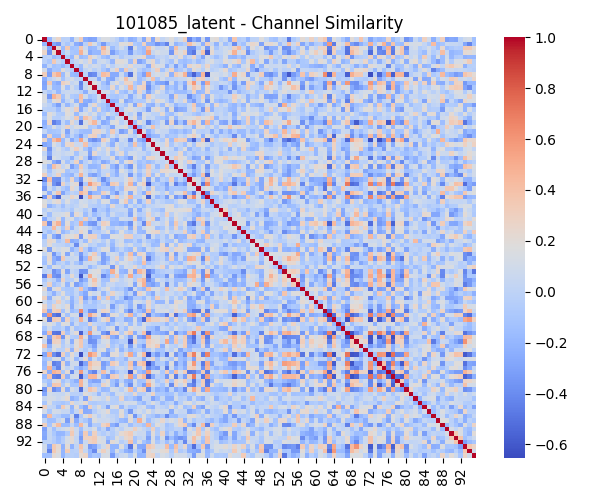}
        \vspace{-5mm}
        \caption{}
    \end{subfigure}
    \vspace{-2mm}
    \caption{
    \textit{(a)-(d)}: The channel-wise similarity matrix from the 1st Scale (\(H \times W \times C\)) to the 4th Scale (\(\nicefrac{H}{8} \times \nicefrac{W}{8} \times 8C\)).
    \textit{(e)}: The channel-wise similarity matrix of (d) after channel-wise projection.
    }
    \label{fig:cross_layer_sim}
    \vspace{-1mm}
\end{figure}

\textbf{SPD Manifold Space Contrastive Learning Leads to More Discriminative Representations.}  
To enhance representation consistency across depth, we introduce a contrastive objective defined over SPD (Symmetric Positive Definite) manifold features. 
We note that the goal here is not to perform full Riemannian optimization along SPD geodesics. 
Instead, we adopt a lightweight formulation that retains the key second‐order structure of covariance matrices while keeping training stable and efficient. 
Strict geodesic contrastive learning typically requires repeated log/exp mappings and matrix decompositions, which incur considerable overhead in large low-level vision models. 
Our approach strikes a practical balance by preserving essential SPD structure before projection. Specifically, in our work,
given shallow features \( F_{\text{shallow}} \in \mathbb{R}^{C_s \times H \times W} \) and latent features \( F_{\text{latent}} \in \mathbb{R}^{C_l \times H' \times W'} \), we first reduce their channel dimensions via \(1 \times 1\) convolutions. The resulting tensors are reshaped into feature matrices \( X_s, X_l \in \mathbb{R}^{C \times N} \) with \( N = H \times W \), and their second-order statistics are computed as:
\begin{equation}
    \mathbf{C}_s = \frac{1}{N - 1} (X_s - \mu_s)(X_s - \mu_s)^\top + \epsilon \mathbf{I}, \quad 
    \mathbf{C}_l = \frac{1}{N' - 1} (X_l - \mu_l)(X_l - \mu_l)^\top + \epsilon \mathbf{I}, 
    \label{eq:spd_cov}
\end{equation}
where \( \mu \) is the mean across spatial dimensions, and \( \epsilon \mathbf{I} \) ensures numerical stability and positive definiteness.
The SPD matrices \( \mathbf{C}_s, \mathbf{C}_l \in \mathbb{R}^{C \times C} \) are vectorized and projected to a contrastive embedding space via shallow 1-layer MLPs:
\begin{equation}
z_s = \operatorname{Norm}(W_s \cdot \text{vec}(\mathbf{C}_s)), \quad 
z_l = \operatorname{Norm}(W_l \cdot \text{vec}(\mathbf{C}_l)),
\label{eq:proj}
\end{equation}
where \( W_s, W_l \) are learnable projection layers, and \( \operatorname{Norm}(\cdot) \) denotes \( \ell_2 \)-normalization.
We then apply an InfoNCE-style contrastive loss to align the shallow and latent embeddings:
\begin{equation}
    \mathcal{L}_{\text{SPD}} = -\log \frac{\exp\left(\operatorname{sim}(z_s, z_l)/\tau\right)}
    {\sum\limits_{z_l'} \exp\left(\operatorname{sim}(z_s, z_l')/\tau\right)},
    \label{eq:contrastive}
\end{equation}
where \( \operatorname{sim}(\cdot,\cdot) \) denotes cosine similarity and \( \tau \) a temperature parameter.
Unlike Euclidean contrastive learning, which views features as flat vectors, our SPD-based method preserves second-order channel dependencies, providing richer structural supervision.
This regularization aligns local and semantic features across depth, enhances discriminability, and \textit{introduces no additional inference cost}.
\section{Experiments}
\label{sec:experiments}
We conduct experiments adhering to the protocols of prior general image restoration works \citep{potlapalli2023promptir,zhang2023ingredient} under 5 settings: 
\textit{(a) 3 Degradations)},
\textit{(b) 5 Degradations)},
\textit{(c) Mixed Degradation},
\textit{(d) Adverse Weather Removal}, and
\textit{(e) Zero-Shot}.
The implementation and experimental details, and the dataset description are provided in the appendix.
Our code, checkpoints, and visual results are available via:
\href{https://github.com/Amazingren/MIRAGE}{\textcolor{cyan}{\texttt{https://github.com/Amazingren/MIRAGE}}}.

\subsection{SOTA Comparison.}
\textbf{3 Degradations.}
We evaluate our method against others listed in Tab.~\ref{tab:exp:3deg}, all trained on three degradations: dehazing, deraining, and denoising. \ourmethod consistently outperforms all the comparison methods, even for those with the assistance of language, multi-task, or prompts. 
Notably, even our \textbf{6M} tiny model outperforms our baseline PromptIR by \textbf{0.71dB} on average. Our 10M small model achieves the best performance across all the metrics, with \textbf{60\%} fewer parameters compared MoCE-IR. 
Compared to DA-RCOT~\citep{tang2025degradation}, which performs contrastive learning over residual feature space, MIRAGE achieves consistently better restoration quality while using substantially fewer parameters (10M vs. 50M). This highlights the efficiency and effectiveness of our SPD-based cross-layer alignment despite its more compact design.

\begin{table}[t]
    \centering
    \scriptsize
    \setlength\tabcolsep{2.5pt}
    \setlength{\extrarowheight}{0.05pt}
    \caption{\textit{Comparison to state-of-the-art on three degradations.} PSNR (dB, $\uparrow$) and SSIM ($\uparrow$) metrics are reported on the full RGB images. \textcolor{tabred}{\textbf{Best}} performances is highlighted. 
    ‘-’ means unreported results.}
    \vspace{-3mm}
    \label{tab:exp:3deg}
    \begin{tabularx}{\textwidth}{p{3.3cm}*{15}{c}}
    \toprule
        \multirow{2}{*}{Method} & \multirow{2}{*}{Venue.} 
     & \multirow{2}{*}{Params.} & \multicolumn{2}{c}{\textit{Dehazing}} & \multicolumn{2}{c}{\textit{Deraining}} & \multicolumn{6}{c}{\textit{Denoising}} 
     & \multicolumn{2}{c}{\multirow{2}{*}{Average}}  \\
     \cmidrule(lr){4-5} \cmidrule(lr){6-7} \cmidrule(lr){8-13} 
     &&& \multicolumn{2}{c}{SOTS} & \multicolumn{2}{c}{Rain100L} & \multicolumn{2}{c}{BSD68\textsubscript{$\sigma$=15}} & \multicolumn{2}{c}{BSD68\textsubscript{$\sigma$=25}} & \multicolumn{2}{c}{BSD68\textsubscript{$\sigma$=50}} &  \\
     \midrule
        \rowcolor{gray!10} BRDNet~\citep{tian2000brdnet} & NN'20 &- & 23.23 & {.895} & 27.42 & {.895} & 32.26 & {.898} & 29.76 & {.836} & 26.34 & {.693} & 27.80 & {.843} \\
        LPNet~\citep{gao2019dynamic} & CVPR'19 &- & 20.84 & {.828} & 24.88 & {.784} & 26.47 & {.778} & 24.77 & {.748} & 21.26 & {.552} & 23.64 & {.738} \\
        \rowcolor{gray!10} FDGAN~\citep{dong2020fdgan}  & AAAI'20&- & 24.71 & {.929} & 29.89 & {.933} & 30.25 & {.910} & 28.81 & {.868} & 26.43 & {.776} & 28.02 & {.883} \\
        DL~\citep{fan2019dl} & TPAMI'19 &2M & 26.92 & {.931} & 32.62 & {.931} & 33.05 & {.914} & 30.41 & {.861} & 26.90 & {.740}  & 29.98 & {.876}\\
        \rowcolor{gray!10} 
        MPRNet~\citep{zamir2021multi} & CVPR'21 & 16M & 25.28 & {.955} & 33.57 & {.954} & 33.54 & {.927} & 30.89 & {.880} & 27.56 & {.779} & 30.17 & {.899} \\
        AirNet~\citep{li2022all} & CVPR'22 & 9M & 27.94 & {.962} & 34.90 & {.967} & 33.92 & {.933} & 31.26 & {.888} & 28.00 & {.797} & 31.20 & {.910} \\
        \rowcolor{gray!10} 
        NDR~\citep{yao2024neural} & TIP'24 & 28M & 25.01 & {.860} & 28.62 & {.848} & 28.72 & {.826} & 27.88 & {.798} & 26.18 & {.720} & 25.01 & {.810} \\
        PromptIR~\citep{potlapalli2023promptir} & NeurIPS'23 & 36M  & 30.58 & .974 & 36.37 &  .972 & 33.98 &  .933 & {31.31} & {.888} & {28.06} &  {.799} & {32.06} &  {.913} \\
        \rowcolor{gray!10} MoCE-IR-S~\citep{zamfir2025moce} & CVPR'25 & 11M  & 30.98 & {.979} & {38.22} &  {.983} & 34.08 &  {.933} & {31.42} & {.888} & {28.16} &  {.798} & {32.57} &  {.916} \\
        AdaIR~\citep{cui2025adair} & ICLR'25 &  29M & 31.06 & .980 & 38.64 & .983 & 34.12 & .935 & 31.45 & .892 & 28.19 & .802 & 32.69 & .918  \\
        \rowcolor{gray!10} MoCE-IR~\citep{zamfir2025moce} & CVPR'25 & 25M & 31.34 &.979 & 38.57 & .984 & 34.11 & .932 &31.45&.888& 28.18&.800 &32.73&.917 \\
        DA-RCOT~\citep{tang2025degradation} & TPAMI'25 & 50M & 31.26 & .977 & 38.36 & .983 & 33.98 & .934 & 31.33 & .890 & 28.10 & .801 & 32.60 & .917 \\
        \midrule
        \rowcolor{green!2}\ourmethod-T (\textit{Ours}) & ICLR'26 & 6M & 31.81 & \sotaa{\textbf{.982}} & 38.44 & .983 & 34.05 & .935 & 31.40 & \sotaa{\textbf{.892}} & 28.14 & .802 & 32.77 & .919 \\
        \rowcolor{green!2}\ourmethod-S (\textit{Ours}) & ICLR'26 & 10M & \sotaa{\textbf{31.86}} & .981 & \sotaa{\textbf{38.94}} & \sotaa{\textbf{.985}} & \sotaa{\textbf{34.12}} & \sotaa{\textbf{.935}} & \sotaa{\textbf{31.46}} & .891 & \sotaa{\textbf{28.19}} & \sotaa{\textbf{.803}} & \sotaa{\textbf{32.91}} & \sotaa{\textbf{.919}} \\ 
        \midrule
        \multicolumn{14}{c}{Methods with the assistance of vision language, multi-task learning, natural language prompts, or multi-modal control} \\
        \midrule 
        \rowcolor{gray!10} DA-CLIP~\citep{luo2023controlling} & ICLR'24& 125M & 29.46 & .963 & 36.28 & .968 &  30.02 & .821 & 24.86 & .585 & 22.29 & .476 & - & - \\
        Art$_{PromptIR}$~\citep{wu2024harmony} &ACM MM'24 & 36M & 30.83 &.979 & 37.94 & .982 &34.06 & .934 & 31.42 & .891 & 28.14 & .801 & 32.49 & .917 \\
        \rowcolor{gray!10} InstructIR-3D~\citep{conde2024high} & ECCV'24 & 16M & 30.22 &.959 & 37.98 & .978 & 34.15& .933 &31.52&.890& 28.30&.804 &32.43&.913 \\
        UniProcessor~\citep{duan2025uniprocessor} &ECCV'24& 1002M & 31.66 & .979 &  38.17 & .982 & 34.08 &.935 &  31.42 & .891 & 28.17 & .803 & 32.70 & .918  \\
        \rowcolor{gray!10}
        VLU-Net~\citep{zeng2025vision} & CVPR'25 & 35M & 30.71 & .980 & 38.93 & .984 & 34.13 & .935 & 31.48 & .892 & 28.23 & .804 & 32.70 & .919 \\
        RamIR~\citep{tang2025ramir} & Applied'25 & 21.7M & 31.29 & .977 & 38.16 & .981 & 34.04 & .931 & 31.61 & .891 & 28.19 & .801 & 32.65 & .916 \\
    \bottomrule
    \end{tabularx}
    \vspace{0mm}
\end{table}

\begin{table}[t]
    \centering
    \scriptsize
    \setlength\tabcolsep{2.2pt}
    \setlength{\extrarowheight}{0.1pt}
    \caption{\textit{Comparison to state-of-the-art on five degradations.} PSNR (dB, $\uparrow$) and SSIM ($\uparrow$) metrics are reported on the full RGB images with $(^\ast)$ denoting general image restorers, others are specialized all-in-one approaches. \textcolor{tabred}{\textbf{Best}} performance is highlighted.}
    \vspace{-3mm}
    \label{tab:exp:5deg}
    \begin{tabularx}{\textwidth}{p{3.6cm}*{15}{c}}
    \toprule
    \multirow{2}{*}{Method} &\multirow{2}{*}{Venue}& \multirow{2}{*}{Params.} 
    & \multicolumn{2}{c}{\textit{Dehazing}} & \multicolumn{2}{c}{\textit{Deraining}} & \multicolumn{2}{c}{\textit{Denoising}} 
    & \multicolumn{2}{c}{\textit{Deblurring}} & \multicolumn{2}{c}{\textit{Low-Light}} & \multicolumn{2}{c}{\multirow{2}{*}{Average}}  \\
    \cmidrule(lr){4-5} \cmidrule(lr){6-7} \cmidrule(lr){8-9} \cmidrule(lr){10-11} \cmidrule(lr){12-13}
    &&& \multicolumn{2}{c}{SOTS} & \multicolumn{2}{c}{Rain100L} & \multicolumn{2}{c}{BSD68\textsubscript{$\sigma$=25}} 
    & \multicolumn{2}{c}{GoPro} & \multicolumn{2}{c}{LOLv1} &  \\
    \midrule
    
    \rowcolor{gray!10} NAFNet$^\ast$~\citep{chen2022simple} & ECCV'22 & 17M & 25.23 & {.939} & 35.56 & {.967} & 31.02 & {.883} & 26.53 & {.808} & 20.49 & {.809} & 27.76 & {.881} \\
    DGUNet$^\ast$~\citep{mou2022deep} & CVPR'22& 17M & 24.78 & {.940} & 36.62 & {.971} & 31.10 & {.883} & 27.25 & {.837} & 21.87 & {.823} & 28.32 & {.891} \\
    \rowcolor{gray!10} SwinIR$^\ast$~\citep{liang2021swinir} &ICCVW'21& 1M & 21.50 & {.891} & 30.78 & {.923} & 30.59 & {.868} & 24.52 & {.773} & 17.81 & {.723} & 25.04 & {.835} \\ 
    Restormer$^\ast$~\citep{zamir2022restormer} &CVPR'22& 26M & 24.09 & {.927} & 34.81 & {.962} & 31.49 & {.884} & 27.22 & {.829} & 20.41 & {.806} & 27.60 & {.881} \\
    \rowcolor{gray!10} MambaIR$^\ast$~\citep{guo2024mambair} &ECCV'24& 27M & 25.81 & .944 & 36.55 & .971 & 31.41 & .884 & 28.61 & .875 & 22.49 & .832 & 28.97 &.901 \\
    \midrule
    DL~\citep{fan2019dl} & TPAMI'19 &2M & 20.54 & {.826} & 21.96 & {.762} & 23.09 & {.745} & 19.86 & {.672} & 19.83 & {.712} & 21.05 & {.743} \\
    \rowcolor{gray!10}Transweather & CVPR'22 & 38M & 21.32 & {.885} & 29.43 & {.905} & 29.00 & {.841} & 25.12 & {.757} & 21.21 & {.792} & 25.22 & {.836} \\ 
    TAPE~\citep{liu2022tape} &ECCV'22& 1M & 22.16 & {.861} & 29.67 & {.904} & 30.18 & {.855} & 24.47 & {.763} & 18.97 & {.621} & 25.09 & {.801} \\
    \rowcolor{gray!10} 
    AirNet~\citep{li2022all} &CVPR'22& 9M & 21.04 & {.884} & 32.98 & {.951} & 30.91 & {.882} & 24.35 & {.781} & 18.18 & {.735} & 25.49 & {.847} \\
    IDR~\citep{zhang2023ingredient} &CVPR'23& 15M & {25.24} & {.943} & {35.63} & {.965} & 31.60 & {.887} & {27.87} & {.846} & 21.34 & {.826} & {28.34} & {.893} \\
    \rowcolor{gray!10}
    PromptIR~\citep{potlapalli2023promptir} &NeurIPS'23& 36M & 30.41 & {.972} & 36.17 & {.970} & 31.20 & {.885} & 27.93 & {.851} & {22.89} & {.829} & 29.72 & {.901} \\
    MoCE-IR-S~\citep{zamfir2025moce} & CVPR'25 & 11M & 31.33 & {.978} & 37.21 & {.978} & {31.25} & {.884} & 28.90 & {.877} & {21.68} & {.851} & 30.08 & {.913} \\
    \rowcolor{gray!10}
    AdaIR~\citep{cui2025adair} & ICLR'25 & 29 & 30.53 & .978 & 38.02 & .981 & 31.35 & .889 & 28.12 & .858 & 23.00 & .845 & 30.20 & .910 \\
    MoCE-IR~\citep{zamfir2025moce} & CVPR'25 & 25M & 30.48 & {.974} & 38.04 & {.982} & {31.34} & {.887} & \textcolor{tabred}{\textbf{30.05}} & \sotaa{\textbf{.899}} & {23.00} & {.852} & 30.58 & \sotaa{\textbf{.919}} \\
    \rowcolor{gray!10} DA-RCOT~\citep{tang2025degradation} & TPAMI'25 & 50M & 30.96 & .975 & 37.87 & .980 & 31.23 & .888 & 28.68 & .872 & 23.25 & .836 & 30.40 & .911 \\
    \midrule
    \rowcolor{green!2}\ourmethod-T \textit{(Ours)} & ICLR'26 & 6M & 31.35 & .979 & 38.24 & .983 & 31.35 & .891 & 27.98 & .850 & 23.11 & .854 & 30.41 & .912 \\
    \rowcolor{green!2}\ourmethod-S \textit{(Ours)} & ICLR'26 & 10M & \sotaa{\textbf{31.45}} & \sotaa{\textbf{.980}} & \sotaa{\textbf{38.92}} & \sotaa{\textbf{.985}} & \sotaa{\textbf{31.41}} & \sotaa{\textbf{.892}} & 28.10 & {.858} & \sotaa{\textbf{23.59}} & \sotaa{\textbf{.858}} & \sotaa{\textbf{30.68}} & {.914} \\
    \midrule
    \multicolumn{14}{c}{Methods with the assistance of natural language prompts or multi-task learning} \\
    \midrule
    \rowcolor{gray!10}
    InstructIR-5D~\citep{conde2024high} &ECCV'24& 16M & 36.84 & .973 & 27.10 & .956 & 31.40 & .887 & 29.40 & .886 & 23.00 & .836 & 29.55 & .908 \\
    Art$_{PromptIR}$~\citep{wu2024harmony} & ACM MM'24& 36M& 29.93 & .908 & 22.09 & .891 & 29.43 & .843 & 25.61 & .776 & 21.99 & .811 & 25.81 & .846 \\
    \rowcolor{gray!10}
    VLU-Net~\citep{zeng2025vision} & CVPR'25 & 35M & 30.84 & .980 & 38.54 & .982 & 31.43 & .891 & 27.46 & .840 & 22.29 & .833 & 30.11 & .905 \\
    RamIR~\citep{tang2025ramir} & Applied'25 & 21.7M & 31.09 & .978 & 37.56 & .979 & 31.44 & .886 & 28.82 & .878 & 22.02 & .828 & 30.18 & .910 \\
    \bottomrule
    \end{tabularx}
    \vspace{0mm}
\end{table}

\textbf{5 Degradations.}
Extending the 3 tasks with deblurring and low-light enhancement~\citep{li2022all,zhang2023ingredient}, we evaluate our \ourmethod's performance in a more challenging 5-degradation setting.
Tab.~\ref{tab:exp:5deg} shows that \ourmethod-S surpasses PromptIR~\citep{potlapalli2023promptir}, MoCE-IR-S~\citep{zamfir2025moce}, AdaIR~\citep{cui2025adair}, and VLU-Net~\citep{zeng2025vision} by \textbf{1.53dB}, \textbf{0.6dB}, \textbf{0.48dB}, and \textbf{0.57dB} on average, with fewer parameters.
Our tiny model (6M) also achieves a second-best average PSNR against MoCE-IR (25M) and surpasses all other methods, including those aided by additional modalities, multi-task learning, or pretraining.

\textbf{Mixed Degradations.}  
\begin{table}[t]
    \centering
    \tiny
    \setlength\tabcolsep{1pt}
    \caption{\textit{Comparison to state-of-the-art on composited degradations.} PSNR (dB, $\uparrow$) and \colorbox{clblue!50}{SSIM ($\uparrow$)} are reported on the full RGB images.
    Our method consistently outperforms even larger models, with favorable results in composited degradation scenarios.}
    \vspace{-3mm}
    \label{tab:exp:cdd11}
    \begin{tabularx}{\textwidth}{p{1.4cm}c*{8}{c}*{10}{c}*{4}{c}cc}
    \toprule
    \multirow{2}{*}{Method} & \multirow{2}{*}{Params.} & \multicolumn{8}{c}{\textit{CDD11-Single}} & \multicolumn{10}{c}{\textit{CDD11-Double}} & \multicolumn{4}{c}{\textit{CDD11-Triple}} & \multicolumn{2}{c}{\multirow{2}{*}{Avg.}}\\
    \cmidrule(lr){3-10} \cmidrule(lr){11-20} \cmidrule(lr){21-24}
    && \multicolumn{2}{c}{Low~(L)} & \multicolumn{2}{c}{Haze~(H)} & \multicolumn{2}{c}{Rain~(R)} & \multicolumn{2}{c}{Snow~(S)}
    & \multicolumn{2}{c}{L+H} & \multicolumn{2}{c}{L+R} & \multicolumn{2}{c}{L+S} & \multicolumn{2}{c}{H+R} & \multicolumn{2}{c}{H+S} 
    &  \multicolumn{2}{c}{L+H+R} &  \multicolumn{2}{c}{L+H+S} \\ 
    \midrule
    \rowcolor{gray!10} 
    AirNet & 9M
    & 24.83&\cc{.778} & 24.21&\cc{.951} & 26.55&\cc{.891} & 26.79&\cc{.919}
    & 23.23&\cc{.779} & 22.82&\cc{.710} & 23.29&\cc{.723} & 22.21&\cc{.868} & 23.29&\cc{.901}
    & 21.80&\cc{.708} & 22.24&\cc{.725} & 23.75&\cc{.814} \\
    PromptIR & 36M
    & 26.32&\cc{.805} & 26.10&\cc{.969} & 31.56&\cc{.946} & 31.53&\cc{.960} 
    & 24.49&\cc{.789} & 25.05&\cc{.771} & 24.51&\cc{.761} & 24.54&\cc{.924} & 23.70&\cc{.925} 
    & 23.74&\cc{.752} & 23.33&\cc{.747} & 25.90&\cc{.850} \\
    \rowcolor{gray!10} 
    WGWSNet & 26M
    & 24.39&\cc{.774} & 27.90&\cc{.982} & 33.15&\cc{.964} & 34.43&\cc{.973} 
    & 24.27&\cc{.800} & 25.06&\cc{.772} & 24.60&\cc{.765} & 27.23&\cc{.955} & 27.65&\cc{.960}  
    & 23.90&\cc{.772} & 23.97&\cc{.771} & 26.96&\cc{.863} \\
    WeatherDiff & 83M
    & 23.58&\cc{.763} & 21.99&\cc{.904} & 24.85&\cc{.885} & 24.80&\cc{.888} 
    & 21.83&\cc{.756} & 22.69&\cc{.730} & 22.12&\cc{.707} & 21.25&\cc{.868} & 21.99&\cc{.868} 
    & 21.23&\cc{.716} & 21.04&\cc{.698} & 22.49&\cc{.799} \\
    \rowcolor{gray!10} 
    OneRestore & 6M
    & 26.48&\cc{{.826}} & 32.52&\cc{.990} & 33.40&\cc{.964} & 34.31&\cc{.973}  
    & 25.79&\cc{{.822}} & 25.58&\cc{.799} & 25.19&\cc{.789} & {29.99}&\cc{.957} & {30.21}&\cc{.964} 
    & 24.78&\cc{.788} & 24.90&\cc{{.791}} & 28.47&\cc{.878} \\
    MoCE-IR & 11M 
    & {27.26}&\cc{.824} & {32.66}&\cc{{.990}} & {34.31}&\cc{{.970}} & {35.91}&\cc{{.980}}
    & {26.24}&\cc{.817} & {26.25}&\cc{{.800}} &{26.04}&\cc{{.793}} & 29.93&\cc{{.964}} & 30.19& \cc{{.970}}
    & {25.41}& \cc{{.789}} &{25.39}&\cc{.790} & {29.05} & \cc{.881} \\
    \midrule
    \rowcolor{green!2} \ourmethod (\textit{ours}) & 6M & 27.13 &\cc{.830} & 32.39 & \cc{.989} & 34.23 & \cc{.969} &35.57 & .978 & 26.04 & \cc{.823} &26.21 & \cc{.807} &26.07 & \cc{.799} & 29.49 & \cc{.962} & 29.72  &\cc{.967} &25.17 & \cc{.793} & 25.41 & \cc{.793} & 28.86 & \cc{.883} \\
    \rowcolor{green!2} \ourmethod (\textit{ours}) & 10M & \sotaa{\textbf{27.41}} & \cc{\sotaa{\textbf{.833}}} & \sotaa{\textbf{33.12}} & \cc{\sotaa{\textbf{.992}}} & \sotaa{\textbf{34.66}} & \cc{\sotaa{\textbf{.971}}} & \sotaa{\textbf{35.98}} & \cc{\sotaa{\textbf{.981}}} & \sotaa{\textbf{26.55}} & \cc{\sotaa{\textbf{.828}}} & \sotaa{\textbf{26.53}} & \cc{\sotaa{\textbf{.810}}} & \sotaa{\textbf{26.33}} & \cc{\sotaa{\textbf{.803}}} &\sotaa{\textbf{30.32}} & \cc{\sotaa{\textbf{.965}}} & \sotaa{\textbf{30.27}} & \cc{\sotaa{\textbf{.969}}} & \sotaa{\textbf{25.59}} & \cc{\sotaa{\textbf{.801}}}& \sotaa{\textbf{25.86}} & \cc{\sotaa{\textbf{.799}}} & \sotaa{\textbf{29.33}} & \cc{\sotaa{\textbf{.887}}}  \\
    \bottomrule
\end{tabularx}
\vspace{0mm}
\end{table}
To better approximate real-world conditions, we extend OneRestore~\citep{guo2024onerestore} to cover both diverse single degradations (rain, haze, snow, low light) and composite cases with multiple degradations per image, yielding eleven distinct restoration settings.
As shown in Tab.~\ref{tab:exp:cdd11}, \ourmethod consistently outperforms leading approaches including AirNet~\citep{li2022all}, PromptIR~\citep{potlapalli2023promptir}, WGWSNet~\citep{ZhuWFYGDQH23}, WeatherDiff~\citep{weather_diffusion}, OneRestore~\citep{guo2024onerestore}, and MoCE-IR~\citep{zamfir2025moce}. Specifically, our Tiny (6M) and Small (10M) models outperform OneRestore~\citep{guo2024onerestore} (6M) by \textbf{0.39 dB} and \textbf{0.86dB} on average. 
Compared to the recent SOTA MoCE-IR~\citep{zamfir2025moce} (11M), our Small model achieves \textbf{0.28dB} higher performance with fewer parameters (10M vs. 11M). These results highlight the effectiveness of our method, particularly for complex mixed degradations.

\begin{table}[t]
    \centering
    \scriptsize
    \setlength\tabcolsep{1.6pt}
    \setlength{\extrarowheight}{0.1pt}
    \newcolumntype{Y}{>{\centering\arraybackslash}X}
    \caption{{Comparisons for \textit{4-task adverse weather removal}}. 
    Missing values are denoted by '--'.}
    \vspace{-3mm}
    \label{tab:exp:ad_weather}
    \begin{tabularx}{\textwidth}{l c *{10}{Y}}
    \toprule
        \multirow{2}{*}{\textbf{Method}} & \multirow{2}{*}{\textbf{Venue}}
        & \multicolumn{2}{c}{\textbf{Snow100K-S}} 
        & \multicolumn{2}{c}{\textbf{Snow100K-L}} 
        & \multicolumn{2}{c}{\textbf{Outdoor-Rain}} 
        & \multicolumn{2}{c}{\textbf{RainDrop}} 
        & \multicolumn{2}{c}{\textbf{Average}} \\
        \cmidrule(r){3-4} \cmidrule(r){5-6} \cmidrule(r){7-8} \cmidrule(r){9-10} \cmidrule(r){11-12}
        & & PSNR & SSIM & PSNR & SSIM & PSNR & SSIM & PSNR & SSIM & PSNR & SSIM \\
        \midrule
        \rowcolor{gray!10}
        All-in-One~\citep{as2020}   & CVPR'20  & -- & -- & 28.33 & .882 & 24.71 & .898 & 31.12 & .927 & 28.05 & .902 \\
        TransWeather~\citep{Transweather}    & CVPR'22   & 32.51 & .934 & 29.31 & .888 & 28.83 & .900 & 30.17 & .916 & 30.20 & .909 \\
        \rowcolor{gray!10}
        Chen~\etal~\citep{ChenHTYDK22} &  CVPR'22  &    34.42 & .947 & 30.22 & .907 & 29.27 & .915 & 31.81 & .931 & 31.43 & .925 \\
        WGWSNet~\citep{ZhuWFYGDQH23}    &  CVPR'23  & 34.31 & .946 & 30.16 & .901 & 29.32 & .921 & 32.38 & .938 & 31.54 & .926 \\
        \rowcolor{gray!10}
        WeatherDiff\(_{64}\)~\citep{weather_diffusion}   & TPAMI'23 & 35.83 & .957 & 30.09 & .904 & 29.64 & .931 & 30.71 & .931 & 31.57 & .931 \\
        WeatherDiff\(_{128}\)~\citep{weather_diffusion}     & TPAMI`23 & 35.02 & .952 & 29.58 & .894 & 29.72 & .922 & 29.66 & .923 & 31.00 & .923 \\
        \rowcolor{gray!10}
        AWRCP~\citep{AWRCP_YeCBSXJYCL23}  & ICCV'23  & 36.92 & .965 & 31.92 & .934 & 31.39 & .933 & 31.93 & .931 & 33.04 & .941 \\
        GridFormer~\citep{Gridformer}    & IJCV'24 & 37.46 & .964 & 31.71 & .923 & 31.87 & .933 & 32.39 & .936 & 33.36 & .939 \\
        \rowcolor{gray!10}
        MPerceiver~\citep{AiHZW024}  & CVPR'24  & 36.23 & .957 & 31.02 & .916 & 31.25 & .925 & \sotaa{\textbf{33.21}} & .929 & 32.93 & .932 \\
        DTPM~\citep{DTPM_0001CCXQL024}  & CVPR'24  & 37.01&  .966 & 30.92 & .917 &  30.99  &  .934 & 32.72 & .944 & 32.91 & .940\\
        \rowcolor{gray!10}
        Histoformer~\citep{Histoformer_SunRGWC24} & ECCV'24 & {37.41} & {.966} & {32.16} & {.926} & {32.08} & {.939} & {33.06} & {.944} & {33.68} & {.944} \\
        \midrule
        \rowcolor{green!2} \ourmethod-S \textit{(Ours)} & ICLR'26 & \sotaa{\textbf{37.97}} & \sotaa{\textbf{.973}} & \sotaa{\textbf{32.33}} & \sotaa{\textbf{.929}} & \sotaa{\textbf{32.82}} & \sotaa{\textbf{.949}} & {{32.78}} & \sotaa{\textbf{.945}} & \sotaa{\textbf{33.98}} & \sotaa{\textbf{.949}}  \\
    \bottomrule
    \end{tabularx}
    \vspace{-3mm}
\end{table}
\textbf{Adverse Weather Removal.}
Following~\citep{Transweather,ZhuWFYGDQH23}, We test our \ourmethod on three challenging deweathering tasks: 
snow removal, 
rain streak and fog removal, 
and raindrop removal.
Tab.~\ref{tab:exp:ad_weather} shows the comparison of our \ourmethod and other \sota methods. 
\ourmethod consistently outperforms existing methods across almost all datasets except PSNR for RainDrop.
The performance gains over multiple weather degradations demonstrate the effectiveness of \ourmethod in handling diverse weather conditions.
Especially, \textbf{0.30dB} improvement on PSNR over Histoformer~\citep{Histoformer_SunRGWC24} and \textbf{1.05dB} improvements over MPerceiver~\citep{AiHZW024}.

\begin{table*}[!t]
\parbox{.4\linewidth}{
    \begin{center}
    \caption{\textit{Zero-Shot} Cross-Domain Underwater Image Enhancement Results.}
    \label{table:ab_colordn_window}
    \vspace{-1mm}
    \setlength{\extrarowheight}{0.15pt}
    \setlength{\tabcolsep}{5pt}
    \scalebox{0.73}{
    \begin{tabular}{l | c@{\hskip 1pt}c }
    \toprule[.1em]
    Method	& PSNR ($\uparrow$) & SSIM ($\uparrow$) 	\\ \midrule
    \rowcolor{gray!10}
    SwinIR~\citep{liang2021swinir}  & 15.31	& .740	\\
    NAFNet~\citep{chu2022nafssr}	& 15.42	& .744 \\	
    \rowcolor{gray!10}
    Restormer~\citep{zamir2022restormer}	& 15.46	& .745 \\
    \midrule
    AirNet~\citep{li2022all}   & 15.46	& .745	\\
    \rowcolor{gray!10}
    IDR~\citep{zhang2023ingredient} & 15.58 & .762\\
    PromptIR~\citep{potlapalli2023promptir} & 15.48 & .748\\ 
    \rowcolor{gray!10}
    MoCE-IR~\citep{zamfir2025moce} & 15.91 & .765 \\
    \midrule
    \rowcolor{green!2} \ourmethod-S \textit{(Ours)} & \sotaa{\textbf{17.29}} & \sotaa{\textbf{.773}} \\
    \bottomrule[.1em]
    \end{tabular}
    }
    \end{center}
}
\hspace{3mm}
\parbox{.53\linewidth}{
    \scriptsize
    \begin{center}
    \caption{\textit{Complexity Analysis.} FLOPs are computed on an image of size $224\times224$ using a NVIDIA Tesla A100 (40G) GPU.}
    \label{table:efficiency}
    \vspace{-1mm}
    \setlength{\extrarowheight}{0.1pt}
    \setlength{\tabcolsep}{5pt}
    \scalebox{0.9}{
    \begin{tabular}{l | c @{\hskip 1pt} c @{\hskip 1pt} c @{\hskip 1pt} c}
    \toprule
    Method	& PSNR ($\uparrow$)	& Memory ($\downarrow$) & Params. ($\downarrow$) & FLOPs ($\downarrow$)\\ 
    \midrule
    \rowcolor{gray!10}
    AirNet~\citep{li2022all}   &31.20 &4829M &  8.93M  & 238G \\
    PromptIR~\citep{potlapalli2023promptir}  & 32.06& 9830M & 35.59M & 132G \\
    \rowcolor{gray!10}
    IDR~\citep{zhang2023ingredient} & - & 4905M & 15.34M & 98G \\
    AdaIR~\citep{cui2025adair} & - & 9740M & 28.79M & 124G  \\
    \rowcolor{gray!10}
    MoCE-IR-S~\citep{zamfir2025moce} &32.51 & 4263M & 11.48M & 37G \\
    MoCE-IR~\citep{zamfir2025moce} & 32.73 & 6654M & 25.35M & 75G \\
    \midrule
    \rowcolor{green!2}\ourmethod-T \textit{(Ours)} & 32.77 & \sotaa{\textbf{3729M}} & \sotaa{\textbf{6.21M}} & \sotaa{\textbf{16G}} \\
    \rowcolor{green!2}\ourmethod-S \textit{(Ours)} & \sotaa{\textbf{32.91}} & 4810M & 9.68M & 27G \\
    \bottomrule
    \end{tabular}
    }
    \end{center}
}
\vspace{-2mm}
\end{table*}
\textbf{Zero-Shot Setting.}  
We evaluate our method’s generalization under a challenging zero-shot setting with real-world underwater images. As shown in Tab.~\ref{table:ab_colordn_window}, \ourmethod-S achieves 17.29 dB and 0.773 SSIM, surpassing MoCE-IR~\citep{zamfir2025moce} by \textbf{+1.38dB} PSNR, while being more compact.
Importantly, our model never sees underwater data during training, yet our adaptive modeling not only fits mixed degradations but also transfers robustly to unseen conditions.
Besides, we also followed the same experimental setting introduced by UniRestore~\cite{{chen2025unirestore}} for the generalization ability evaluation. 
Meanwhile, the real-world evaluation presented in Tab.~\ref{tab:exp:udc_zero_shot} shows that MIRAGE generalizes reliably to real-world, camera-captured degradations.

\textbf{Efficiency Comparison.}  
Tab.~\ref{table:efficiency} compares PSNR, memory, parameters, and FLOPs. Our Tiny model (\ourmethod-T), with only 6.21M parameters and 16G FLOPs, delivers the best efficiency–performance trade-off, outperforming all prior methods, including larger models like PromptIR~\citep{potlapalli2023promptir} and MoCE-IR-S~\citep{zamfir2025moce}. It surpasses MoCE-IR-S by \textbf{+0.26 dB} while using less than half the computation, and even our Small variant (\ourmethod-S) exceeds full MoCE-IR in both PSNR (\textbf{+0.18dB}) and FLOPs (27G vs. 75G). These results confirm that our design achieves strong restoration quality without compromising efficiency.

\textbf{Visual Comparison.}  
\ourmethod effectively restores fine structural details and reliably suppresses subtle visual artifacts across diverse and unseen degradations (Fig.~\ref{fig:teaser} and appendix).

\subsection{Ablation Analysis \& Discussion}
\textbf{Components ablation.} Tab.~\ref{tab:ablation} shows starting from an attention-only setting (32.23 dB, 19.89M), 
we progressively integrate each module while reducing complexity.
\begin{wraptable}{r}{0.55\textwidth}
    \centering
    \scriptsize
    \setlength\tabcolsep{5pt}
    \setlength{\extrarowheight}{0.05pt}
    \vspace{0mm}
    \caption{\textit{Ablation Study} of \ourmethod-T under the 3-Degradation Setting with Tiny model.}
    \label{tab:ablation}
    \vspace{-2mm}
    \scalebox{0.99}{
    \begin{tabular}{lccc}
    \toprule
    \multirow{2}{*}{Ablaton} & \multirow{2}{*}{Params.} &  \multicolumn{2}{c}{Results} \\
    \cmidrule(lr){3-4} && PSNR (dB, $\uparrow$) & SSIM($\downarrow$) \\
    \midrule 
    \rowcolor{gray!10}
    att-only \textit{(Ours)} & 19.89 M & 32.23 (\sotab{-0.54}) & .912 \\
    \textit{w/o} DynamicConv & 9.43 M & 32.21 (\sotab{-0.56}) & .911 \\
    \rowcolor{gray!10}
    \textit{w/o} C-MLP & 7.01 M & 32.39 (\sotab{-0.38}) & .913 \\
    \textit{w/o} Fusion (\ie Cat()-Only) & 5.71 M & 32.57 (\sotab{-0.20}) & .914 \\
    \midrule
    \rowcolor{gray!10}
    \textit{w/o} CL \& SPD & 5.80M & 32.63 (\sotab{-0.14}) & .916\\
    \textit{w/o} SPD (CL Euclidean) & 6.10M & 32.53 (\sotab{-0.24}) & .914 \\
    \midrule
    \rowcolor{green!2}\ourmethod-T \textit{(Full) }& 6.21M & \sotaa{32.77} & \sotaa{.919} \\
    \bottomrule
    \end{tabular}
    }
    \vspace{0mm}
\end{wraptable}
Removing the dynamic convolution branch (\textit{w/o DynamicConv}) causes a 0.56 dB drop, indicating its importance for local spatial modeling. 
The channel-wise MLP (\textit{w/o C-MLP}) also plays a critical role, with a 0.38 dB performance loss. 
Naive concatenation (\textit{w/o Fusion}) leads to a further 0.20 dB drop, confirming that explicit feature integration is more effective.
On the regularization side, removing contrastive learning (\textit{w/o CL \& SPD}) or replacing SPD with Euclidean alignment degrades performance by 0.14 dB and 0.24 dB, indicating that structure-agnostic contrastive learning can misguide optimization, while manifold-aware alignment provides consistent benefits.
Overall, each component contributes to the final performance. Our full model offers the best balance between accuracy and efficiency with only 6.21M parameters and 32.77 dB PSNR.

\begin{wrapfigure}{r}{0.40\textwidth}
    \centering
    \vspace{-3mm}
    \includegraphics[width=0.88\linewidth]{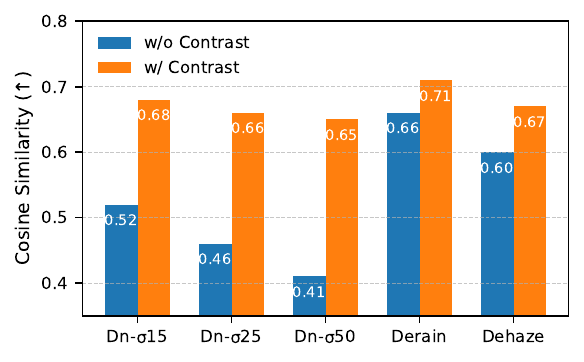}
    \vspace{-4mm}
    \caption{Shallow–latent cosine similarity across degradations. Contrastive alignment improves feature correlation.}
    \vspace{-2mm}
    \label{fig:cosine-sim-plot}
\end{wrapfigure}
\textbf{Why shallow–latent Contrastive Alignment Matters.}
Different degradations rely on different feature levels: denoising and deraining benefit from shallow, texture-rich features, while dehazing and low-light enhancement require deeper semantic features; deblurring needs both. 
This heterogeneity makes unified modeling challenging. 
We therefore introduce contrastive alignment between shallow and latent stages to encourage semantic coordination. 
When shallow features dominate (\eg, denoising), alignment guides latent features to be more task-relevant; when latent features dominate (\eg, dehazing), shallow features inherit semantic consistency~\citep{bertasius2015high}. 
Fig.~\ref{fig:cosine-sim-plot} validates that contrastive alignment improves shallow–latent correlation, validating its necessity for cross-degradation generalization.

\begin{wrapfigure}{r}{0.41\textwidth}
  \vspace{-8pt}
  \centering
  \includegraphics[width=0.96\linewidth]{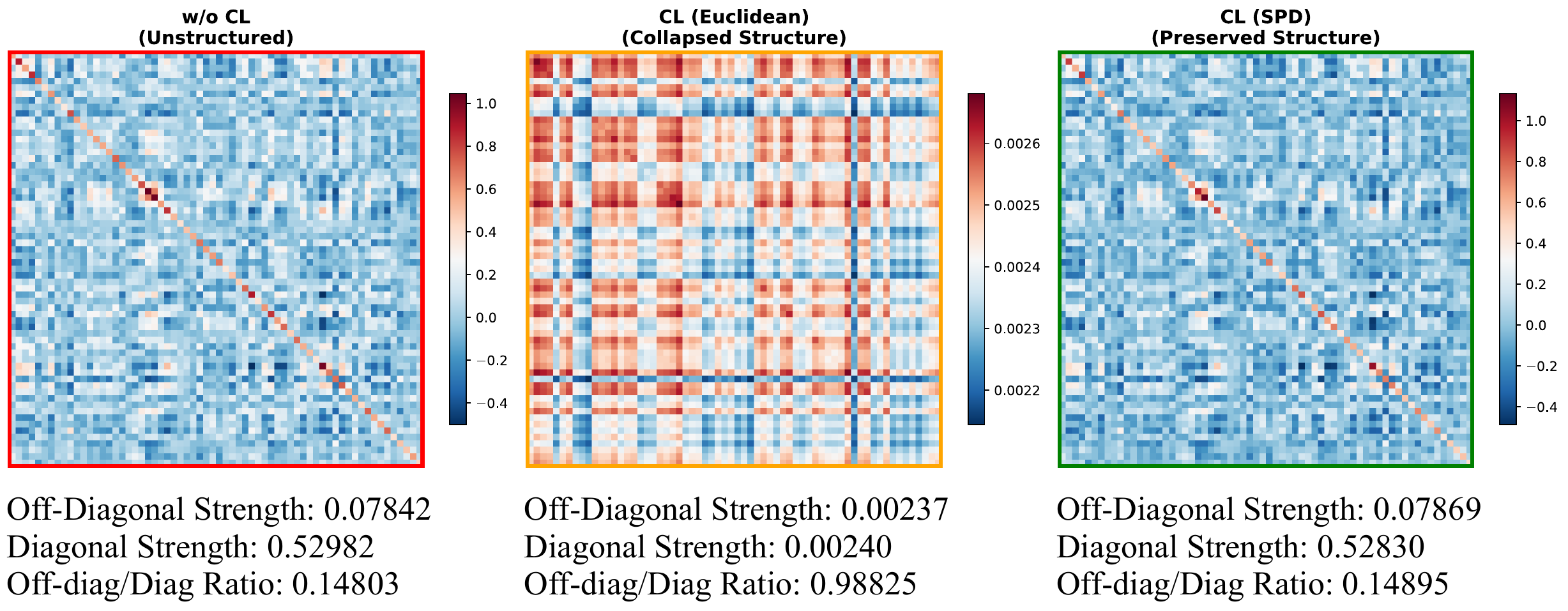}%
  \vspace{-6pt}
  \caption{Shallow–latent similarity under three settings:
  (a) w/o CL (unstructured; off-diag $0.0784$, ratio $0.148$),
  (b) Euclidean CL (collapsed; off-diag $\approx 0.0024$, ratio $0.99$),
  (c) SPD CL (preserved; off-diag $0.0787$, ratio $0.149$).
  }
  \label{fig:heatmap}
  \vspace{-6pt}
\end{wrapfigure}
\textbf{Why Euclidean Fails and Why SPD Works? (Deraining Case Study)}  
Euclidean contrastive learning collapses shallow–latent alignment by enforcing indiscriminate similarity, reducing both diagonal and off-diagonal terms to trivial constants, and erasing task cues. SPD, by aligning covariance matrices on a Riemannian manifold, preserves second-order dependencies and guides updates along meaningful directions. In the deraining case (Figure~\ref{fig:heatmap}), Euclidean CL degenerates into near-constant similarity (off-diag $0.00237$, ratio $0.99$), while SPD maintains diagonal dominance and non-trivial off-diagonal structure ($0.0787$, ratio $0.149$), producing coherent patterns.

\section{Conclusion}
\label{sec:conclusion}
We presented \ourmethod, an efficient framework for degradation-agnostic image restoration that achieves a favorable balance between robustness and efficiency. Through channel-wise functional decomposition, the model repurposes redundant capacity into convolution-, attention-, and MLP-based branches, enabling complementary modeling of local textures, global context, and channel-wise statistics. To further enhance cross-degradation generalization, we introduced manifold regularization, aligning shallow and latent features in the SPD manifold space for more consistent and discriminative representations. Extensive experiments across diverse degradations, including mixed and unseen scenarios, demonstrate that \ourmethod achieves state-of-the-art performance. 
Inspired by the metaphor of a mirage, \ie, revealing the hidden reality beneath visual distortions, our framework learns degradation-agnostic representations by balancing global, local, and channel-wise information, providing a scalable foundation for future research in degradation-agnostic IR.

\subsubsection*{Acknowledgments}
This work was partially supported by the FIS project GUIDANCE (Debugging Computer Vision Models via Controlled Cross-modal Generation) (No. FIS2023-03251).

\section*{Ethics Statement}
Our work focuses on general-purpose image restoration, aiming to improve efficiency and robustness across diverse degradation types. The intended positive impact includes deployment in low-resource or safety-critical scenarios such as mobile photography, remote sensing, medical imaging, and environmental monitoring. At the same time, we recognize that improved restoration techniques could be misused for deceptive content editing or large-scale surveillance. We encourage responsible use of our method and provide our models and code with appropriate licenses and documentation to support transparency and ethical adoption. No personally identifiable or sensitive data were used in this research.  

\section*{Reproducibility Statement}
We aim to ensure reproducibility and transparency of our results. The MIRAGE framework is implemented in PyTorch with standard training protocols and evaluation metrics. Detailed descriptions of the architecture, training settings, datasets, and baselines are provided in the main paper and supplementary material. 
Upon acceptance, we will release the full code, pretrained models, and instructions for reproducing all reported results, including ablation studies and comparisons. 
Random seeds and hardware details are also documented to facilitate faithful replication.

\bibliography{iclr2026_conference}

@String(PAMI = {IEEE Trans. Pattern Anal. Mach. Intell.})

@String(IJCV = {Int. J. Comput. Vis.})

@String(CVPR= {IEEE Conf. Comput. Vis. Pattern Recog.})

@String(ICCV= {Int. Conf. Comput. Vis.})

@String(ECCV= {Eur. Conf. Comput. Vis.})

@String(BMVC= {Brit. Mach. Vis. Conf.})

@String(TIP  = {IEEE Trans. Image Process.})

@String(ICASSP=	{ICASSP})

@String(ICLR = {Int. Conf. Learn. Represent.})

@String(AAAI = {AAAI})

@String(CVPRW= {IEEE Conf. Comput. Vis. Pattern Recog. Worksh.})

@String(PAMI  = {IEEE TPAMI})

@String(IJCV  = {IJCV})

@String(CVPR  = {CVPR})

@String(ICCV  = {ICCV})

@String(ECCV  = {ECCV})

@String(BMVC  =	{BMVC})

@String(TIP   = {IEEE TIP})

@String(TCSVT = {IEEE TCSVT})

@String(ICLR  = {ICLR})

@String(CVPRW= {CVPRW})

@String(ICCVW= {ICCVW})

@String(ICML= {ICML})

@article{richardson1972bayesian,
  title={Bayesian-based iterative method of image restoration},
  author={Richardson, William Hadley},
  journal={Journal of the Optical Society of America},
  volume={62},
  number={1},
  pages={55--59},
  year={1972},
  publisher={Optica Publishing Group}
}

@article{banham1997digital,
  title={Digital image restoration},
  author={Banham, Mark R and Katsaggelos, Aggelos K},
  journal={IEEE Signal Processing Magazine},
  volume={14},
  number={2},
  pages={24--41},
  year={1997},
  publisher={IEEE}
}

@inproceedings{li2023lsdir,
  title={{LSDIR}: A large scale dataset for image restoration},
  author={Li, Yawei and Zhang, Kai and Liang, Jingyun and Cao, Jiezhang and Liu, Ce and Gong, Rui and Zhang, Yulun and Tang, Hao and Liu, Yun and Demandolx, Denis and Ranjan, Rakesh and Timofte, Radu and Van Gool, Luc},
  booktitle=CVPRW,
  pages={1775--1787},
  year={2023}
}

@inproceedings{zamfir2024details,
  title={See More Details: Efficient Image Super-Resolution by Experts Mining}, 
  author={Eduard Zamfir and Zongwei Wu and Nancy Mehta and Yulun Zhang and Radu Timofte},
  booktitle={ICML},
  year={2024},
  organization={PMLR}
}

@inproceedings{lim2017enhanced,
  title={Enhanced deep residual networks for single image super-resolution},
  author={Lim, Bee and Son, Sanghyun and Kim, Heewon and Nah, Seungjun and Lee, Kyoung Mu},
  booktitle=CVPRW,
  pages={1132--1140},
  year={2017}
}

@inproceedings{lai2017deep,
  title={Deep laplacian pyramid networks for fast and accurate super-resolution},
  author={Lai, Wei-Sheng and Huang, Jia-Bin and Ahuja, Narendra and Yang, Ming-Hsuan},
  booktitle=CVPR,
  pages={624--632},
  year={2017}
}

@inproceedings{liang2021swinir,
  title={{SwinIR}: Image restoration using {Swin} transformer},
  author={Liang, Jingyun and Cao, Jiezhang and Sun, Guolei and Zhang, Kai and Van Gool, Luc and Timofte, Radu},
  booktitle=ICCVW,
  pages={1833--1844},
  year={2021}
}

@inproceedings{chen2021learning,
  title={Learning continuous image representation with local implicit image function},
  author={Chen, Yinbo and Liu, Sifei and Wang, Xiaolong},
  booktitle=CVPR,
  pages={8628--8638},
  year={2021}
}

@inproceedings{li2023efficient,
  title={Efficient and explicit modelling of image hierarchies for image restoration},
  author={Li, Yawei and Fan, Yuchen and Xiang, Xiaoyu and Demandolx, Denis and Ranjan, Rakesh and Timofte, Radu and Van Gool, Luc},
  booktitle=CVPR,
  pages={18278--18289},
  year={2023}
}

@article{zhang2024transcending,
  title={Transcending the Limit of Local Window: Advanced Super-Resolution Transformer with Adaptive Token Dictionary},
  author={Zhang, Leheng and Li, Yawei and Zhou, Xingyu and Zhao, Xiaorui and Gu, Shuhang},
  journal={arXiv preprint arXiv:2401.08209},
  year={2024}
}

@inproceedings{gao2023implicit,
  title={Implicit diffusion models for continuous super-resolution},
  author={Gao, Sicheng and Liu, Xuhui and Zeng, Bohan and Xu, Sheng and Li, Yanjing and Luo, Xiaoyan and Liu, Jianzhuang and Zhen, Xiantong and Zhang, Baochang},
  booktitle=CVPR,
  pages={10021--10030},
  year={2023}
}

@article{wang2022zero,
  title={Zero-Shot Image Restoration Using Denoising Diffusion Null-Space Model},
  author={Wang, Yinhuai and Yu, Jiwen and Zhang, Jian},
  journal=ICLR,
  year={2023}
}

@article{luo2023image,
  title={Image restoration with mean-reverting stochastic differential equations},
  author={Luo, Ziwei and Gustafsson, Fredrik K and Zhao, Zheng and Sj{\"o}lund, Jens and Sch{\"o}n, Thomas B},
  journal={arXiv preprint arXiv:2301.11699},
  year={2023}
}

@article{yue2023resshift,
  title={{ResShift}: Efficient Diffusion Model for Image Super-resolution by Residual Shifting},
  author={Yue, Zongsheng and Wang, Jianyi and Loy, Chen Change},
  journal={arXiv preprint arXiv:2307.12348},
  year={2023}
}

@inproceedings{zhao2024denoising,
  title={Denoising Diffusion Probabilistic Models for Action-Conditioned 3D Motion Generation},
  author={Zhao, Mengyi and Liu, Mengyuan and Ren, Bin and Dai, Shuling and Sebe, Nicu},
  booktitle={ICASSP},
  pages={4225--4229},
  year={2024},
  organization={IEEE}
}

@inproceedings{dong2015compression,
  title={Compression artifacts reduction by a deep convolutional network},
  author={Dong, Chao and Deng, Yubin and Loy, Chen Change and Tang, Xiaoou},
  booktitle=ICCV,
  pages={576--584},
  year={2015}
}

@inproceedings{zhang2017learning,
  title={Learning deep CNN denoiser prior for image restoration},
  author={Zhang, Kai and Zuo, Wangmeng and Gu, Shuhang and Zhang, Lei},
  booktitle=CVPR,
  pages={3929--3938},
  year={2017}
}

@article{zhang2017beyond,
  title={Beyond a gaussian denoiser: Residual learning of deep cnn for image denoising},
  author={Zhang, Kai and Zuo, Wangmeng and Chen, Yunjin and Meng, Deyu and Zhang, Lei},
  journal=TIP,
  volume={26},
  number={7},
  pages={3142--3155},
  year={2017},
  publisher={IEEE}
}

@inproceedings{wang2018recovering,
  title={Recovering realistic texture in image super-resolution by deep spatial feature transform},
  author={Wang, Xintao and Yu, Ke and Dong, Chao and Loy, Chen Change},
  booktitle=CVPR,
  pages={606--615},
  year={2018}
}

@inproceedings{tu2022maxim,
  title={{MAXIM}: Multi-axis mlp for image processing},
  author={Tu, Zhengzhong and Talebi, Hossein and Zhang, Han and Yang, Feng and Milanfar, Peyman and Bovik, Alan and Li, Yinxiao},
  booktitle=CVPR,
  pages={5769--5780},
  year={2022}
}

@inproceedings{guo2024mambair,
    title={MambaIR: A Simple Baseline for Image Restoration with State-Space Model},
    author={Guo, Hang and Li, Jinmin and Dai, Tao and Ouyang, Zhihao and Ren, Xudong and Xia, Shu-Tao},
    booktitle={ECCV},
    year={2024}
}

@article{zhu2024vision,
  title={Vision mamba: Efficient visual representation learning with bidirectional state space model},
  author={Zhu, Lianghui and Liao, Bencheng and Zhang, Qian and Wang, Xinlong and Liu, Wenyu and Wang, Xinggang},
  journal={arXiv preprint arXiv:2401.09417},
  year={2024}
}

@article{mamba,
  title={Mamba: Linear-Time Sequence Modeling with Selective State Spaces},
  author={Gu, Albert and Dao, Tri},
  journal={arXiv preprint arXiv:2312.00752},
  year={2023}
}

@inproceedings{mamba2,
  title={Transformers are {SSM}s: Generalized Models and Efficient Algorithms Through Structured State Space Duality},
  author={Dao, Tri and Gu, Albert},
  booktitle={ICML},
  year={2024}
}

@inproceedings{zamir2022restormer,
  title={Restormer: Efficient transformer for high-resolution image restoration},
  author={Zamir, Syed Waqas and Arora, Aditya and Khan, Salman and Hayat, Munawar and Khan, Fahad Shahbaz and Yang, Ming-Hsuan},
  booktitle=CVPR,
  pages={5728--5739},
  year={2022}
}

@inproceedings{ren2023masked,
  title={Masked Jigsaw Puzzle: A Versatile Position Embedding for Vision Transformers},
  author={Ren, Bin and Liu, Yahui and Song, Yue and Bi, Wei and Cucchiara, Rita and Sebe, Nicu and Wang, Wei},
  booktitle=CVPR,
  pages={20382--20391},
  year={2023}
}

@inproceedings{dosovitskiy2020image,
  title={An image is worth 16x16 words: Transformers for image recognition at scale},
  author={Dosovitskiy, Alexey and Beyer, Lucas and Kolesnikov, Alexander and Weissenborn, Dirk and Zhai, Xiaohua and Unterthiner, Thomas and Dehghani, Mostafa and Minderer, Matthias and Heigold, Georg and Gelly, Sylvain and others},
  booktitle=ICLR,
  year={2020}
}

@article{zhang2019residual,
  title={Residual non-local attention networks for image restoration},
  author={Zhang, Yulun and Li, Kunpeng and Li, Kai and Zhong, Bineng and Fu, Yun},
  journal={arXiv preprint arXiv:1903.10082},
  year={2019}
}

@inproceedings{wu2021contrastive,
  title={Contrastive learning for compact single image dehazing},
  author={Wu, Haiyan and Qu, Yanyun and Lin, Shaohui and Zhou, Jian and Qiao, Ruizhi and Zhang, Zhizhong and Xie, Yuan and Ma, Lizhuang},
  booktitle={CVPR},
  pages={10551--10560},
  year={2021}
}

@inproceedings{jiang2020multi,
  title={Multi-scale progressive fusion network for single image deraining},
  author={Jiang, Kui and Wang, Zhongyuan and Yi, Peng and Chen, Chen and Huang, Baojin and Luo, Yimin and Ma, Jiayi and Jiang, Junjun},
  booktitle={CVPR},
  pages={8346--8355},
  year={2020}
}

@inproceedings{kong2023efficient,
  title={Efficient frequency domain-based transformers for high-quality image deblurring},
  author={Kong, Lingshun and Dong, Jiangxin and Ge, Jianjun and Li, Mingqiang and Pan, Jinshan},
  booktitle={CVPR},
  pages={5886--5895},
  year={2023}
}

@inproceedings{li2022all,
  title={All-in-one image restoration for unknown corruption},
  author={Li, Boyun and Liu, Xiao and Hu, Peng and Wu, Zhongqin and Lv, Jiancheng and Peng, Xi},
  booktitle={CVPR},
  pages={17452--17462},
  year={2022}
}

@article{potlapalli2023promptir,
  title={Promptir: Prompting for all-in-one image restoration},
  author={Potlapalli, Vaishnav and Zamir, Syed Waqas and Khan, Salman H and Shahbaz Khan, Fahad},
  journal={NeurIPS},
  volume={36},
  year={2024}
}

@inproceedings{zhang2023ingredient,
  title={Ingredient-oriented multi-degradation learning for image restoration},
  author={Zhang, Jinghao and Huang, Jie and Yao, Mingde and Yang, Zizheng and Yu, Hu and Zhou, Man and Zhao, Feng},
  booktitle={CVPR},
  pages={5825--5835},
  year={2023}
}

@article{wang2023promptrestorer,
  title={Promptrestorer: A prompting image restoration method with degradation perception},
  author={Wang, Cong and Pan, Jinshan and Wang, Wei and Dong, Jiangxin and Wang, Mengzhu and Ju, Yakun and Chen, Junyang},
  journal={NeurIPS},
  volume={36},
  pages={8898--8912},
  year={2023}
}

@article{li2023prompt,
  title={Prompt-in-prompt learning for universal image restoration},
  author={Li, Zilong and Lei, Yiming and Ma, Chenglong and Zhang, Junping and Shan, Hongming},
  journal={arXiv preprint arXiv:2312.05038},
  year={2023}
}

@article{dudhane2024dynamic,
  title={Dynamic Pre-training: Towards Efficient and Scalable All-in-One Image Restoration},
  author={Dudhane, Akshay and Thawakar, Omkar and Zamir, Syed Waqas and Khan, Salman and Khan, Fahad Shahbaz and Yang, Ming-Hsuan},
  journal={arXiv preprint arXiv:2404.02154},
  year={2024}
}

@article{tian2000brdnet,
    title = {Image denoising using deep CNN with batch renormalization},
    journal = {Neural Networks},
    year = {2020},
    author = {Chunwei Tian and Yong Xu and Wangmeng Zuo},
}

@inproceedings{gao2019dynamic,
  title={Dynamic scene deblurring with parameter selective sharing and nested skip connections},
  author={Gao, Hongyun and Tao, Xin and Shen, Xiaoyong and Jia, Jiaya},
  booktitle=CVPR,
  year={2019}
}

@inproceedings{dong2020fdgan,
  title={FD-GAN: Generative adversarial networks with fusion-discriminator for single image dehazing},
  author={Dong, Yu and Liu, Yihao and Zhang, He and Chen, Shifeng and Qiao, Yu},
  booktitle={AAAI},
  year={2020}
}

@article{fan2019dl,
  title={A general decoupled learning framework for parameterized image operators},
  author={Fan, Qingnan and Chen, Dongdong and Yuan, Lu and Hua, Gang and Yu, Nenghai and Chen, Baoquan},
  journal={TPAMI},
  volume={43},
  number={1},
  pages={33--47},
  year={2019},
  publisher={IEEE}
}

@article{yao2024neural,
  title={Neural degradation representation learning for all-in-one image restoration},
  author={Yao, Mingde and Xu, Ruikang and Guan, Yuanshen and Huang, Jie and Xiong, Zhiwei},
  journal={TIP},
  year={2024},
  publisher={IEEE}
}

@inproceedings{luo2023controlling,
  title={Controlling vision-language models for universal image restoration},
  author={Luo, Ziwei and Gustafsson, Fredrik K and Zhao, Zheng and Sj{\"o}lund, Jens and Sch{\"o}n, Thomas B},
  booktitle={ICLR},
  year={2024}
}

@inproceedings{conde2024high,
  title={InstructIR: High-Quality Image Restoration Following Human Instructions},
  author={Conde, Marcos V and Geigle, Gregor and Timofte, Radu},
  booktitle    = {ECCV},
  year={2024}
}

@inproceedings{duan2025uniprocessor,
  title={Uniprocessor: a text-induced unified low-level image processor},
  author={Duan, Huiyu and Min, Xiongkuo and Wu, Sijing and Shen, Wei and Zhai, Guangtao},
  booktitle={ECCV},
  pages={180--199},
  year={2025},
  organization={Springer}
}

@inproceedings{wu2024harmony,
  title={Harmony in Diversity: Improving All-in-One Image Restoration via Multi-Task Collaboration},
  author={Wu, Gang and Jiang, Junjun and Jiang, Kui and Liu, Xianming},
  booktitle={Proceedings of the 32nd ACM International Conference on Multimedia},
  pages={6015--6023},
  year={2024}
}

@inproceedings{zamir2021multi,
  title={Multi-stage progressive image restoration},
  author={Zamir, Syed Waqas and Arora, Aditya and Khan, Salman and Hayat, Munawar and Khan, Fahad Shahbaz and Yang, Ming-Hsuan and Shao, Ling},
  booktitle=CVPR,
  pages={14821--14831},
  year={2021}
}

@inproceedings{cui2025adair,
    title={Ada{IR}: Adaptive All-in-One Image Restoration via Frequency Mining and Modulation},
    author={Yuning Cui and Syed Waqas Zamir and Salman Khan and Alois Knoll and Mubarak Shah and Fahad Shahbaz Khan},
    booktitle={ICLR},
    year={2025}
}

@inproceedings{chen2022simple,
  title={Simple baselines for image restoration},
  author={Chen, Liangyu and Chu, Xiaojie and Zhang, Xiangyu and Sun, Jian},
  booktitle=ECCV,
  pages={17--33},
  year={2022}
}

@inproceedings{mou2022deep,
  title={Deep generalized unfolding networks for image restoration},
  author={Mou, Chong and Wang, Qian and Zhang, Jian},
  booktitle={CVPR},
  pages={17399--17410},
  year={2022}
}

@inproceedings{Transweather,
    author    = {Jeya Maria Jose Valanarasu and Rajeev Yasarla and Vishal M. Patel},
    booktitle = {CVPR},
    title     = {TransWeather: Transformer-based Restoration of Images Degraded by Adverse Weather Conditions},
    pages     = {2343--2353},
    year      = {2022},
}

@inproceedings{liu2022tape,
  title={Tape: Task-agnostic prior embedding for image restoration},
  author={Liu, Lin and Xie, Lingxi and Zhang, Xiaopeng and Yuan, Shanxin and Chen, Xiangyu and Zhou, Wengang and Li, Houqiang and Tian, Qi},
booktitle = {ECCV},
  year={2022},
}

@article{arbelaez2010contour,
  title={Contour detection and hierarchical image segmentation},
  author={Arbelaez, Pablo and Maire, Michael and Fowlkes, Charless and Malik, Jitendra},
  journal=PAMI,
  volume={33},
  number={5},
  pages={898--916},
  year={2010},
  publisher={IEEE}
}

@article{ma2016waterloo,
  title={Waterloo exploration database: New challenges for image quality assessment models},
  author={Ma, Kede and Duanmu, Zhengfang and Wu, Qingbo and Wang, Zhou and Yong, Hongwei and Li, Hongliang and Zhang, Lei},
  journal=TIP,
  volume={26},
  number={2},
  pages={1004--1016},
  year={2016},
  publisher={IEEE}
}

@inproceedings{martin2001database,
  title={A database of human segmented natural images and its application to evaluating segmentation algorithms and measuring ecological statistics},
  author={Martin, David and Fowlkes, Charless and Tal, Doron and Malik, Jitendra},
  booktitle=ICCV,
  pages={416--423},
  year={2001}
}

@inproceedings{huang2015single,
  title={Single image super-resolution from transformed self-exemplars},
  author={Huang, Jia-Bin and Singh, Abhishek and Ahuja, Narendra},
  booktitle=CVPR,
  pages={5197--5206},
  year={2015}
}

@inproceedings{yang2020learning,
  title={Learning texture transformer network for image super-resolution},
  author={Yang, Fuzhi and Yang, Huan and Fu, Jianlong and Lu, Hongtao and Guo, Baining},
  booktitle={CVPR},
  pages={5791--5800},
  year={2020}
}

@article{li2018benchmarking,
  title={Benchmarking single-image dehazing and beyond},
  author={Li, Boyi and Ren, Wenqi and Fu, Dengpan and Tao, Dacheng and Feng, Dan and Zeng, Wenjun and Wang, Zhangyang},
  journal={TIP},
  volume={28},
  number={1},
  pages={492--505},
  year={2018},
  publisher={IEEE}
}

@inproceedings{nah2017deep,
  title={Deep multi-scale convolutional neural network for dynamic scene deblurring},
  author={Nah, Seungjun and Hyun Kim, Tae and Mu Lee, Kyoung},
  booktitle=CVPR,
  pages={3883--3891},
  year={2017}
}

@article{wei2018deep,
  title={Deep retinex decomposition for low-light enhancement},
  author={Wei, Chen and Wang, Wenjing and Yang, Wenhan and Liu, Jiaying},
  journal={arXiv preprint arXiv:1808.04560},
  year={2018}
}

@inproceedings{guo2024onerestore,
  title={OneRestore: A Universal Restoration Framework for Composite Degradation},
  author={Guo, Yu and Gao, Yuan and Lu, Yuxu and Liu, Ryan Wen and He, Shengfeng},
  booktitle={ECCV},
  year={2024}
}

@inproceedings{kingma2015adam,
  title={Adam: A Method for Stochastic Optimization},
  author={Kingma, Diederik P and Ba, Jimmy},
  booktitle=ICLR,
  year={2015}
}

@inproceedings{zeng2025vision,
  title={Vision-Language Gradient Descent-driven All-in-One Deep Unfolding Networks},
  author={Zeng, Haijin and Wang, Xiangming and Chen, Yongyong and Su, Jingyong and Liu, Jie},
  booktitle={CVPR},
  year={2025}
}

@inproceedings{zamfir2025moce,
  title={Complexity Experts are Task-Discriminative Learners for Any Image Restoration}, 
  author={Eduard Zamfir and Zongwei Wu and Nancy Mehta and Yuedong Tan and Danda Pani Paudel and Yulun Zhang and Radu Timofte},
  booktitle={CVPR},
  year={2025},
}

@inproceedings{raindrop,
  author       = {Rui Qian and
                  Robby T. Tan and
                  Wenhan Yang and
                  Jiajun Su and
                  Jiaying Liu},
  title        = {Attentive Generative Adversarial Network for Raindrop Removal From
                  a Single Image},
  booktitle    = {CVPR},
  pages        = {2482--2491},
  year         = {2018}
}

@inproceedings{as2020,
    title={All in one bad weather removal using architectural search},
    author={Li, Ruoteng and Tan, Robby T and Cheong, Loong-Fah},
    booktitle={CVPR},
    pages={3175--3185},
    year={2020}
}

@inproceedings{ChenHTYDK22,
    author       = {Wei{-}Ting Chen and
                  Zhi{-}Kai Huang and
                  Cheng{-}Che Tsai and
                  Hao{-}Hsiang Yang and
                  Jian{-}Jiun Ding and
                  Sy{-}Yen Kuo},
    title        = {Learning Multiple Adverse Weather Removal via Two-stage Knowledge
                  Learning and Multi-contrastive Regularization: Toward a Unified Model},
    booktitle    = {CVPR},
    pages        = {17632--17641},
    year         = {2022}
}

@inproceedings{ZhuWFYGDQH23,
    author    = {Zhu, Yurui and Wang, Tianyu and Fu, Xueyang and Yang, Xuanyu and Guo, Xin and Dai, Jifeng and Qiao, Yu and Hu, Xiaowei},
    booktitle = {CVPR},
    title     = {Learning weather-general and weather-specific features for image restoration under multiple adverse weather conditions},
    pages     = {21747--21758},
    year      = {2023},
}

@article{weather_diffusion,
    author  = {Ozan {\"{O}}zdenizci and Robert Legenstein},
    title   = {Restoring Vision in Adverse Weather Conditions With Patch-Based Denoising Diffusion Models},
    number  = {8},
    pages   = {10346--10357},
    volume  = {45},
    journal = {TPAMI},
    year    = {2023},
}

@inproceedings{AWRCP_YeCBSXJYCL23,
  author       = {Tian Ye and
                  Sixiang Chen and
                  Jinbin Bai and
                  Jun Shi and
                  Chenghao Xue and
                  Jingxia Jiang and
                  Junjie Yin and
                  Erkang Chen and
                  Yun Liu},
  title        = {Adverse Weather Removal with Codebook Priors},
  booktitle    = {ICCV},
  pages        = {12619--12630},
  year         = {2023}
}

@article{Gridformer,
  title={Gridformer: Residual dense transformer with grid structure for image restoration in adverse weather conditions},
  author={Wang, Tao and Zhang, Kaihao and Shao, Ziqian and Luo, Wenhan and Stenger, Bjorn and Lu, Tong and Kim, Tae-Kyun and Liu, Wei and Li, Hongdong},
  journal={IJCV},
  volume={132},
  number={10},
  pages={4541--4563},
  year={2024},
  publisher={Springer}
}

@inproceedings{AiHZW024,
    author       = {Yuang Ai and
                  Huaibo Huang and
                  Xiaoqiang Zhou and
                  Jiexiang Wang and
                  Ran He},
    title        = {Multimodal Prompt Perceiver: Empower Adaptiveness, Generalizability
                  and Fidelity for All-in-One Image Restoration},
    booktitle    = {CVPR},
    pages        = {25432--25444},
    year         = {2024}
}

@inproceedings{DTPM_0001CCXQL024,
    author       = {Tian Ye and
                  Sixiang Chen and
                  Wenhao Chai and
                  Zhaohu Xing and
                  Jing Qin and
                  Ge Lin and
                  Lei Zhu},
    title        = {Learning Diffusion Texture Priors for Image Restoration},
    booktitle    = {{CVPR}},
    pages        = {2524--2534},
    year         = {2024}
}

@inproceedings{Histoformer_SunRGWC24,
    author       = {Shangquan Sun and
                  Wenqi Ren and
                  Xinwei Gao and
                  Rui Wang and
                  Xiaochun Cao},
    title        = {Restoring Images in Adverse Weather Conditions via Histogram Transformer},
    booktitle    = {ECCV},
    volume       = {15080},
    pages        = {111--129},
    year         = {2024}
}

@article{cai2016dehazenet,
  title={Dehazenet: An end-to-end system for single image haze removal},
  author={Cai, Bolun and Xu, Xiangmin and Jia, Kui and Qing, Chunmei and Tao, Dacheng},
  journal=TIP,
  volume={25},
  number={11},
  pages={5187--5198},
  year={2016},
  publisher={IEEE}
}

@inproceedings{li2017aod,
  title={Aod-net: All-in-one dehazing network},
  author={Li, Boyi and Peng, Xiulian and Wang, Zhangyang and Xu, Jizheng and Feng, Dan},
  booktitle={ICCV},
  pages={4770--4778},
  year={2017}
}

@inproceedings{qu2019enhanced,
  title={Enhanced pix2pix dehazing network},
  author={Qu, Yanyun and Chen, Yizi and Huang, Jingying and Xie, Yuan},
  booktitle={CVPR},
  pages={8160--8168},
  year={2019}
}

@inproceedings{ren2016single,
  title={Single image dehazing via multi-scale convolutional neural networks},
  author={Ren, Wenqi and Liu, Si and Zhang, Hua and Pan, Jinshan and Cao, Xiaochun and Yang, Ming-Hsuan},
  booktitle=ECCV,
  pages={154--169},
  year={2016}
}

@inproceedings{zhang2018density,
  title={Density-aware single image de-raining using a multi-stream dense network},
  author={Zhang, He and Patel, Vishal M},
  booktitle={CVPR},
  pages={695--704},
  year={2018}
}

@inproceedings{yasarla2019uncertainty,
  title={Uncertainty guided multi-scale residual learning-using a cycle spinning cnn for single image de-raining},
  author={Yasarla, Rajeev and Patel, Vishal M},
  booktitle={CVPR},
  pages={8405--8414},
  year={2019}
}

@inproceedings{wei2019semi,
  title={Semi-supervised transfer learning for image rain removal},
  author={Wei, Wei and Meng, Deyu and Zhao, Qian and Xu, Zongben and Wu, Ying},
  booktitle={CVPR},
  pages={3877--3886},
  year={2019}
}

@article{zhang2018ffdnet,
  title={{FFDNet}: Toward a fast and flexible solution for CNN-based image denoising},
  author={Zhang, Kai and Zuo, Wangmeng and Zhang, Lei},
  journal=TIP,
  volume={27},
  number={9},
  pages={4608--4622},
  year={2018},
  publisher={IEEE}
}

@article{ren2024sharing,
  title={Sharing key semantics in transformer makes efficient image restoration},
  author={Ren, Bin and Li, Yawei and Liang, Jingyun and Ranjan, Rakesh and Liu, Mengyuan and Cucchiara, Rita and Gool, Luc V and Yang, Ming-Hsuan and Sebe, Nicu},
  journal={NeurIPS},
  volume={37},
  pages={7427--7463},
  year={2024}
}

@inproceedings{lee2015deeply,
  title={Deeply-supervised nets},
  author={Lee, Chen-Yu and Xie, Saining and Gallagher, Patrick and Zhang, Zhengyou and Tu, Zhuowen},
  booktitle={Artificial intelligence and statistics},
  pages={562--570},
  year={2015},
  organization={Pmlr}
}

@inproceedings{hu2025universal,
    title={Universal Image Restoration Pre-training via Degradation Classification},
    author={Hu JiaKui and Zhengjian Yao and Jin Lujia and Lu Yanye},
    booktitle={ICLR},
    year={2025},
}

@InProceedings{chu2022nafssr,
    author    = {Chu, Xiaojie and Chen, Liangyu and Yu, Wenqing},
    title     = {NAFSSR: Stereo Image Super-Resolution Using NAFNet},
    booktitle = {CVPRWorkshops},
    month     = {June},
    year      = {2022},
    pages     = {1239-1248}
}

@article{xie2025mat,
  title={MAT: Multi-range attention transformer for efficient image super-resolution},
  author={Xie, Chengxing and Zhang, Xiaoming and Li, Linze and Fu, Yuqian and Gong, Biao and Li, Tianrui and Zhang, Kai},
  journal={TCSVT},
  year={2025},
  publisher={IEEE}
}

@article{nguyen2022improving,
  title={Improving transformer with an admixture of attention heads},
  author={Nguyen, Tan and Nguyen, Tam and Do, Hai and Nguyen, Khai and Saragadam, Vishwanath and Pham, Minh and Nguyen, Khuong Duy and Ho, Nhat and Osher, Stanley},
  journal={NeurIPS},
  volume={35},
  pages={27937--27952},
  year={2022}
}

@inproceedings{nguyen2022improvingkeys,
  title={Improving transformers with probabilistic attention keys},
  author={Nguyen, Tam Minh and Nguyen, Tan Minh and Le, Dung DD and Nguyen, Duy Khuong and Tran, Viet-Anh and Baraniuk, Richard and Ho, Nhat and Osher, Stanley},
  booktitle={ICML},
  pages={16595--16621},
  year={2022},
  organization={PMLR}
}

@inproceedings{xiao2024improving,
  title={Improving Transformers with Dynamically Composable Multi-Head Attention},
  author={Xiao, Da and Meng, Qingye and Li, Shengping and Yuan, Xingyuan},
  booktitle={ICML},
  pages={54300--54318},
  year={2024},
  organization={PMLR}
}

@article{wang2022improved,
  title={Improved transformer with multi-head dense collaboration},
  author={Wang, Huadong and Shen, Xin and Tu, Mei and Zhuang, Yimeng and Liu, Zhiyuan},
  journal={TASLP},
  volume={30},
  pages={2754--2767},
  year={2022},
  publisher={IEEE}
}

@inproceedings{venkataramanan2023skip,
  title={Skip-attention: Improving vision transformers by paying less attention},
  author={Venkataramanan, Shashanka and Ghodrati, Amir and Asano, Yuki M and Porikli, Fatih and Habibian, Amirhossein},
  booktitle=ICLR,
  year={2024}
}

@inproceedings{liu2023spatio,
  title={Spatio-Temporal Graph Diffusion for Text-Driven Human Motion Generation.},
  author={Liu, Chang and Zhao, Mengyi and Ren, Bin and Liu, Mengyuan and Sebe, Nicu and others},
  booktitle={BMVC},
  pages={722--729},
  year={2023}
}

@inproceedings{dong2021attention,
  title={Attention is not all you need: Pure attention loses rank doubly exponentially with depth},
  author={Dong, Yihe and Cordonnier, Jean-Baptiste and Loukas, Andreas},
  booktitle={ICML},
  pages={2793--2803},
  year={2021},
  organization={PMLR}
}

@article{liu2018desnownet,
  title={Desnownet: Context-aware deep network for snow removal},
  author={Liu, Yun-Fu and Jaw, Da-Wei and Huang, Shih-Chia and Hwang, Jenq-Neng},
  journal={TIP},
  volume={27},
  number={6},
  pages={3064--3073},
  year={2018},
  publisher={IEEE}
}

@inproceedings{zheng2024learning,
  title={Learning modality-agnostic representation for semantic segmentation from any modalities},
  author={Zheng, Xu and Lyu, Yuanhuiyi and Wang, Lin},
  booktitle={ECCV},
  pages={146--165},
  year={2024},
  organization={Springer}
}

@article{brodermann2025cafuser,
  title={CAFuser: Condition-Aware Multimodal Fusion for Robust Semantic Perception of Driving Scenes},
  author={Br{\"o}dermann, Tim and Sakaridis, Christos and Fu, Yuqian and Van Gool, Luc},
  journal={RAL},
  year={2025},
  publisher={IEEE}
}

@article{li2025fractal,
  title={Fractal-IR: A Unified Framework for Efficient and Scalable Image Restoration},
  author={Li, Yawei and Ren, Bin and Liang, Jingyun and Ranjan, Rakesh and Liu, Mengyuan and Sebe, Nicu and Yang, Ming-Hsuan and Benini, Luca},
  journal={arXiv preprint arXiv:2503.17825},
  year={2025}
}

@inproceedings{qian2018attentive,
  title={Attentive generative adversarial network for raindrop removal from a single image},
  author={Qian, Rui and Tan, Robby T and Yang, Wenhan and Su, Jiajun and Liu, Jiaying},
  booktitle={CVPR},
  pages={2482--2491},
  year={2018}
}

@inproceedings{li2019heavy,
  title={Heavy rain image restoration: Integrating physics model and conditional adversarial learning},
  author={Li, Ruoteng and Cheong, Loong-Fah and Tan, Robby T},
  booktitle={CVPR},
  pages={1633--1642},
  year={2019}
}

@article{li2019underwater,
  title={An underwater image enhancement benchmark dataset and beyond},
  author={Li, Chongyi and Guo, Chunle and Ren, Wenqi and Cong, Runmin and Hou, Junhui and Kwong, Sam and Tao, Dacheng},
  journal={TIP},
  volume={29},
  pages={4376--4389},
  year={2019},
  publisher={IEEE}
}

@article{jiang2024survey,
  title={A survey on all-in-one image restoration: Taxonomy, evaluation and future trends},
  author={Jiang, Junjun and Zuo, Zengyuan and Wu, Gang and Jiang, Kui and Liu, Xianming},
  journal={IEEE TPAMI},
  year={2025}
}

@article{tang2025degradation,
  title={Degradation-aware residual-conditioned optimal transport for unified image restoration},
  author={Tang, Xiaole and Gu, Xiang and He, Xiaoyi and Hu, Xin and Sun, Jian},
  journal={TPAMI},
  year={2025},
  publisher={IEEE}
}

@inproceedings{tian2025degradation,
  title={Degradation-Aware Feature Perturbation for All-in-One Image Restoration},
  author={Tian, Xiangpeng and Liao, Xiangyu and Liu, Xiao and Li, Meng and Ren, Chao},
  booktitle={CVPR},
  pages={28165--28175},
  year={2025}
}

@article{he2025diffusion,
  title={Diffusion models in low-level vision: A survey},
  author={He, Chunming and Shen, Yuqi and Fang, Chengyu and Xiao, Fengyang and Tang, Longxiang and Zhang, Yulun and Zuo, Wangmeng and Guo, Zhenhua and Li, Xiu},
  journal={TPAMI},
  year={2025},
  publisher={IEEE}
}

@article{zhai2023comprehensive,
  title={A comprehensive review of deep learning-based real-world image restoration},
  author={Zhai, Lujun and Wang, Yonghui and Cui, Suxia and Zhou, Yu},
  journal={Access},
  volume={11},
  pages={21049--21067},
  year={2023},
  publisher={IEEE}
}

@inproceedings{bertasius2015high,
  title={High-for-low and low-for-high: Efficient boundary detection from deep object features and its applications to high-level vision},
  author={Bertasius, Gedas and Shi, Jianbo and Torresani, Lorenzo},
  booktitle={ICCV},
  pages={504--512},
  year={2015}
}

@misc{openai2023gpt4,
      title={GPT-4 Technical Report}, 
      author={OpenAI},
      year={2023},
      eprint={2303.08774},
      archivePrefix={arXiv},
      primaryClass={cs.CL}
}

@misc{ChatGPT,
    author={OpenAI},
    title={Introducing chatgpt},
    howpublished={https://openai.com/blog/chatgpt},
    year={2022}
}

@article{tang2025ramir,
  title={RamIR: Reasoning and action prompting with Mamba for all-in-one image restoration},
  author={Tang, Aiqiang and Wu, Yan and Zhang, Yuwei},
  journal={Applied Intelligence},
  volume={55},
  number={4},
  pages={258},
  year={2025},
  publisher={Springer}
}

@inproceedings{chen2022unpaired,
  title={Unpaired deep image deraining using dual contrastive learning},
  author={Chen, Xiang and Pan, Jinshan and Jiang, Kui and Li, Yufeng and Huang, Yufeng and Kong, Caihua and Dai, Longgang and Fan, Zhentao},
  booktitle={CVPR},
  pages={2017--2026},
  year={2022}
}

@article{chen2025multi,
  title={Multi-modal degradation feature learning for unified image restoration based on contrastive learning},
  author={Chen, Lei and Xiong, Qingbo and Zhang, Wei and Liang, Xiaoli and Gan, Zhihua and Li, Liqiang and He, Xin},
  journal={Neurocomputing},
  volume={616},
  pages={128955},
  year={2025},
  publisher={Elsevier}
}

@article{zhang2025perceive,
  title={Perceive-ir: Learning to perceive degradation better for all-in-one image restoration},
  author={Zhang, Xu and Ma, Jiaqi and Wang, Guoli and Zhang, Qian and Zhang, Huan and Zhang, Lefei},
  journal={TIP},
  year={2025},
  publisher={IEEE}
}

@article{ren2026any,
  title={Any image restoration via efficient spatial-frequency degradation adaptation},
  author={Ren, Bin and Zamfir, Eduard and Wu, Zongwei and Li, Yawei and Li, Yidi and Paudel, Danda Pani and Timofte, Radu and Yang, Ming-Hsuan and Van Gool, Luc and Sebe, Nicu},
  journal={TMLR},
  year={2026}
}

@article{shi2025vmambair,
  title={Vmambair: Visual state space model for image restoration},
  author={Shi, Yuan and Xia, Bin and Jin, Xiaoyu and Wang, Xing and Zhao, Tianyu and Xia, Xin and Xiao, Xuefeng and Yang, Wenming},
  journal={TSCVT},
  year={2025},
  publisher={IEEE}
}

@inproceedings{chen2025unirestore,
  title={Unirestore: Unified perceptual and task-oriented image restoration model using diffusion prior},
  author={Chen, I and Chen, Wei-Ting and Liu, Yu-Wei and Chiang, Yuan-Chun and Kuo, Sy-Yen and Yang, Ming-Hsuan and others},
  booktitle={CVPR},
  pages={17969--17979},
  year={2025}
}

@inproceedings{zheng2024selective,
  title={Selective hourglass mapping for universal image restoration based on diffusion model},
  author={Zheng, Dian and Wu, Xiao-Ming and Yang, Shuzhou and Zhang, Jian and Hu, Jian-Fang and Zheng, Wei-Shi},
  booktitle={CVPR},
  pages={25445--25455},
  year={2024}
}

@inproceedings{zhou2021image,
  title={Image restoration for under-display camera},
  author={Zhou, Yuqian and Ren, David and Emerton, Neil and Lim, Sehoon and Large, Timothy},
  booktitle={Proceedings of the ieee/cvf conference on computer vision and pattern recognition},
  pages={9179--9188},
  year={2021}
}
\bibliographystyle{iclr2026_conference}

\appendix

\setcounter{section}{0}
\setcounter{figure}{0}    
\setcounter{table}{0}   
\setcounter{algorithm}{0}
\setcounter{page}{1}
\setcounter{equation}{0}

\renewcommand{\thetable}{\Alph{table}}
\renewcommand{\thefigure}{\Alph{figure}}
\renewcommand{\thealgorithm}{\Alph{algorithm}}
\renewcommand{\thesection}{\Alph{section}}
\renewcommand{\theequation}{\Alph{equation}}

\section{Experimental Protocols}
\label{suppsec:exp_setup}
\subsection{Datasets}
\textbf{3 Degradation Datasets.}
For both the All-in-One and single-task settings, we follow the evaluation protocols established in prior works~\cite{li2022all,potlapalli2023promptir,zamfir2025moce}, utilizing the following datasets:
For image denoising in the single-task setting, we combine the BSD400~\cite{arbelaez2010contour} and WED~\cite{ma2016waterloo} datasets, and corrupt the images with Gaussian noise at levels $\sigma \in \{15, 25, 50\}$. 
BSD400 contains 400 training images, while WED includes 4,744 images. 
We evaluate the denoising performance on BSD68~\cite{martin2001database} and Urban100~\cite{huang2015single}.
For single-task deraining, we use Rain100L~\cite{yang2020learning}, which provides 200 clean/rainy image pairs for training and 100 pairs for testing.
For single-task dehazing, we adopt the SOTS dataset~\cite{li2018benchmarking}, consisting of 72,135 training images and 500 testing images.
Under the All-in-One setting, we train a unified model on the combined set of the aforementioned training datasets for 120 epochs and directly test it across all three restoration tasks.

\textbf{5 Degradation Datasets.}
The 5-degradation setting is built upon the 3-degradation setting, with two additional tasks included: deblurring and low-light enhancement. 
For deblurring, we adopt the GoPro dataset~\cite{nah2017deep}, which contains 2,103 training images and 1,111 testing images. 
For low-light enhancement, we use the LOL-v1 dataset~\cite{wei2018deep}, consisting of 485 training images and 15 testing images. 
Note that for the denoising task under the 5-degradation setting, we report results using Gaussian noise with $\sigma = 25$. The training takes 130 epochs.

\textbf{Composited Degradation Datasets.}
Regarding the composite degradation setting, we use the CDD11 dataset~\cite{guo2024onerestore}. CDD11 consists of 1,183 training images for:
\textit{(i) 4 kinds of single-degradation types:} haze (H), low-light (L), rain (R), and snow (S);
\textit{(ii) 5 kinds of double-degradation types:} low-light + haze (l+h), low-light+rain (L+R), low-light + snow (L+S), haze + rain (H+R), and haze + snow (H+S).
\textit{(iii) 2 kinds of Triple-degradation type:} low-light + haze + rain (L+H+R), and low-light + haze + snow (L+H+S).
We train our method for 170 epochs (fewer than 200 epochs than MoCE-IR~\cite{zamfir2025moce}), and we keep all other settings unchanged.

\textbf{Adverse Weather Removal Datasets.}
For the deweathering tasks, we follow the experimental setups used in TransWeather~\cite{Transweather} and WGWSNet~\cite{ZhuWFYGDQH23}, evaluating the performance of our approach on multiple synthetic datasets. 
We assess the capability of \ourmethod across three challenging tasks: snow removal, rain streak and fog removal, and raindrop removal. 
The training set, referred to as ``AllWeather'', is composed of images from the Snow100K~\cite{liu2018desnownet}, Raindrop~\cite{qian2018attentive}, and Outdoor-Rain~\cite{li2019heavy} datasets.
For testing, we evaluate our model on the following subsets: Snow100K-S (16,611 images), Snow100K-L (16,801 images), Outdoor-Rain (750 images), and Raindrop (249 images). Same as Histoformer~\cite{Histoformer_SunRGWC24}, we train \ourmethod on ``AllWeather'' with 300,000 iterations.

\textbf{Zero-Shot Underwater Image Enhancement Dataset.}
For the zero-shot underwater image enhancement setting, we follow the evaluation protocol of DCPT~\cite{hu2025universal} by directly applying our model, trained under the 5-degradation setting, on the UIEB dataset~\cite{li2019underwater} without any finetuning. 
UIEB consists of two subsets: 890 raw underwater images with corresponding high-quality reference images, and 60 challenging underwater images. 
We evaluate our zero-shot performance on the 890-image subset with available reference images.

\subsection{Implementation Details}
\noindent \textbf{Implementation Details.}
Our \ourmethod framework is designed to be end-to-end trainable, removing the need for multi-stage optimization of individual components. 
The architecture adopts a robust 4-level encoder-decoder structure, with a varying number of Mixed Degradation Attention Blocks (MDAB) at each level—specifically $[3, 5, 5, 7]$ from highest to lowest resolution in the Tiny variant. 
Following prior works~\cite{potlapalli2023promptir,zamfir2025moce}, we train the model for 120 epochs with a batch size of 32 in both the 3-Degradation All-in-One and single-task settings. 
The optimization uses a combination of $L_{1}$ and Fourier loss, optimized with Adam~\cite{kingma2015adam} (initial learning rate of $\num{2e-4}$, $\beta_1=0.9$, $\beta_2=0.999$) and a cosine decay schedule. 
During training, we apply random cropping to $128{\times}128$ patches, along with horizontal and vertical flipping as data augmentation. 
All experiments are conducted on a single NVIDIA H200 GPU (140 GB). Memory usage is approximately 42~GB for the Tiny (\ie, \ourmethod-T) model and 56~GB for the Small model (\ie, \ourmethod-S).

\textbf{Model Scaling.} We propose two scaled variants of our \ourmethod, namely Tiny (\ourmethod-T) and Small (\ourmethod-S). As detailed in Tab.~\ref{tab:supp:model_details}, these variants differ in terms of the number of MDAB blocks across scales, the input embedding dimension, the FFN expansion factor, and the number of refinement blocks.

\begin{table}[!t]
    \centering
    \scriptsize
    \caption{The details our the tiny and small version of our \ourmethod. FLOPs are computed on an image of size 224 × 224 using a NVIDIA Tesla A100 (40G) GPU.}
    \setlength{\extrarowheight}{0.1pt}
    \setlength\tabcolsep{16pt} 
    \label{tab:supp:model_details}
    \vspace{2mm}
    \scalebox{1.0}{
    \begin{tabular}{l|c|c}
    \toprule
     & \ourmethod-T & \ourmethod-S \\ 
     \midrule
    The Number of the MDAB crosses 4 scales & [3, 5, 5, 7] & [3, 5, 5, 7] \\
    The Input Embedding Dimension & 24 & 30 \\
    The FFN Expansion Factor & 2 & 2\\
    The Number of the Refinement Blocks & 2 & 3\\
    \midrule
    Params. ($\downarrow$) & 6.21M & 9.68 M \\
    FLOPs ($\downarrow$) & 16 G & 27 G \\
    \bottomrule
    \end{tabular}
    }
\end{table}

\subsection{Optimization Objectives}
The overall optimization objective of our approach is defined as:
\begin{equation}
\mathcal{L}_{\text{total}} = \mathcal{L}_{\text{1}} + \lambda_{fre} \times \mathcal{L}_{\text{Fourier}} + \lambda_{ctrs} \times \mathcal{L}_{\text{SPD}}.
\end{equation}
Here, $\mathcal{L}_{\text{Fourier}}$ denotes the real-valued Fourier loss computed between the restored image and the ground-truth image, and $\mathcal{L}_{\text{SPD}}$ represents our proposed contrastive learning objective in the SPD (Symmetric Positive Definite) space.

Specifically, we adopt an $\ell_1$ loss that adopted in IR tasks~\cite{potlapalli2023promptir,zamfir2025moce,li2022all,ren2026any,cui2025adair,ren2024sharing,li2025fractal}, defined as $\mathcal{L}_{\text{1}} = | \hat{x} - x |_1$, to enforce pixel-wise similarity between the restored image $\hat{x}$ and the ground-truth image $x$.
$\mathcal{L}_{\text{Fourier}}$, as utilized in MoCE-IR~\cite{zamfir2025moce,cui2025adair}, to enhance frequency-domain consistency, the real-valued Fourier loss, is defined as:
\begin{equation}
\mathcal{L}_{\text{Fourier}} = \left\| \mathcal{F}_{\text{real}}(\hat{x}) - \mathcal{F}_{\text{real}}(x) \right\|_1 + \left\| \mathcal{F}_{\text{imag}}(\hat{x}) - \mathcal{F}_{\text{imag}}(x) \right\|_1,
\end{equation}
where $\hat{x}$ and $x$ denote the restored and ground-truth images, respectively. $\mathcal{F}_{\text{real}}(\cdot)$ and $\mathcal{F}_{\text{imag}}(\cdot)$ represent the real and imaginary parts of the 2D real-input FFT (\ie, $\mathrm{rfft2}$). 
The final loss is computed as the $\ell_1$ distance between the real and imaginary components of the predicted and target frequency spectra. 
Same as MoCE-IR~\cite{zamfir2025moce}, $\lambda_{fre}$ is set to 0.1 throughout our experiments.
Meanwhile, the $\mathcal{L}_{\text{SPD}}$ is defined as in Eq. 3-5 of our main manuscript. More ablation studies regarding the proposed $\mathcal{L}_{\text{SPD}}$ are provided in Sec.~\ref{subsec:supp:spd_loss}. The temperature parameter \( \tau \) of the proposed \(\mathcal{L}_{\text{SPD}}\) is set to 0.1 throughout all the experiments.

\section{Preliminaries on SPD-based Feature Statistics}
\label{suppsec:preliminary}
This section provides a brief background on the concepts involved in our cross-layer alignment strategy.  
The intention is to supply intuitive context—rather than additional derivations—for second-order feature statistics, the SPD structure, and depth-asymmetric representations used in Sec.~\ref{subsec:spd_ctrs}.

\noindent\textbf{Second-order feature statistics.}
Raw activations capture local appearance, but the way channels vary together often reveals more stable information about degradations.  
For a feature matrix $X \in \mathbb{R}^{C \times N}$, the covariance
\[
\mathbf{C} = \frac{1}{N-1}(X-\mu)(X-\mu)^{\top}
\]
summarizes inter-channel relationships.  
Diagonal entries reflect each channel’s variability, while off-diagonal entries describe redundancy and dependence patterns.  
These structures differ consistently across layers and degradations (Fig.~\ref{fig:cross_layer_sim}; Appendix Fig.~\ref{fig:supp:sim_vis_3deg}), making covariance a compact and informative descriptor.

\noindent\textbf{SPD property of covariance matrices.}
Covariance matrices are symmetric and positive definite by construction and therefore lie in the SPD set.  
This structure encodes meaningful geometric information: eigenvalues represent correlation strengths, and the matrix as a whole can be interpreted as a “shape’’ in channel space.  
Preserving this structure is important—direct Euclidean operations may flatten or distort correlation patterns, an effect also reflected in the collapse observed with Euclidean contrastive learning (Fig.~\ref{fig:heatmap}).

\noindent\textbf{Representing SPD matrices for comparison.}
To compare covariance matrices within a contrastive objective, we vectorize $\mathbf{C}$ and apply a learnable projection.  
This retains second-order relationships while mapping them to an embedding space suitable for contrastive learning.  
Compared to raw feature vectors, covariance embeddings emphasize structural organization and therefore provide a more stable alignment signal.

\noindent\textbf{Depth-asymmetric representations.}
Shallow and latent features naturally exhibit different statistical behavior: shallow layers respond strongly to local degradations and show pronounced redundancy, while deeper layers become more decorrelated and semantically aggregated.  
Their covariance matrices reflect these differences in a consistent way across degradations, making shallow–latent pairs complementary views of the same signal and a natural target for alignment.

\noindent\textbf{Intuition behind SPD-based alignment.}
Aligning covariance-based SPD embeddings focuses on how channels interact, rather than on individual activation values.  
This yields supervision that is less sensitive to local noise and more reflective of the underlying representation structure.  
Encouraging shallow and latent features to share similar second-order statistics stabilizes the shared feature space required for diverse degradations.

Overall, covariance provides a compact view of channel interactions, the SPD structure preserves meaningful second-order relations, and depth-asymmetric covariance patterns naturally motivate the alignment strategy formalized in Sec.~\ref{subsec:spd_ctrs}.

\section{More Method Details \& Supplementary Experiments}
\label{suppsec:add_exp}
\subsection{1 Deg. Comparison}
\begin{table}[!t]
    \centering
    \scriptsize
    \setlength\tabcolsep{2pt}    
    \setlength{\extrarowheight}{0.2pt}
    \caption{\textit{Comparison to state-of-the-art for single degradations.} PSNR (dB, $\uparrow$) and SSIM ($\uparrow$) metrics are reported on the full RGB images. \textcolor{tabred}{\textbf{Best}} performance is highlighted. Our method excels over prior works. 
    }    
    \vspace{-2mm}
    \label{tab:exp:single}
    \scalebox{0.84}{
    \begin{subtable}[l]{0.3\textwidth}
        \subcaption{\textit{Dehazing}}
        \vspace{-1mm}
        \begin{tabularx}{\textwidth}{p{1.8cm}*{4}{c}}
        \toprule
        Method & Params. &\multicolumn{2}{c}{SOTS}\\
        \midrule
        \rowcolor{gray!10}DehazeNet& - & 22.46 & .851\\
        MSCNN & - &22.06&.908 \\
        \rowcolor{gray!10}AODNet & - &20.29&.877\\
        EPDN & - &22.57 & .863\\
        \rowcolor{gray!10}FDGAN & - &23.15 & .921 \\
        \midrule
        AirNet & 9M &23.18 & .900 \\
        \rowcolor{gray!10}PromptIR& 36M & 
        31.31 &
        {.973} \\
        \midrule 
        \rowcolor{green!2}\ourmethod(\textit{Ours}) &
        \textcolor{tabred}{\textbf{6M}} & {31.46} & {.977} \\
        \rowcolor{green!2}\ourmethod(\textit{Ours}) & 10M & \textcolor{tabred}{\textbf{31.53}} & \textcolor{tabred}{\textbf{.980}} \\
        \bottomrule
        \end{tabularx}
    \end{subtable}%
    \hfill
    \hspace{2mm}
    \begin{subtable}[l]{0.3\textwidth}
        \subcaption{\textit{Deraining}}
        \vspace{-1mm}
        \begin{tabularx}{\textwidth}{X*{4}{c}}
        \toprule
        Method & Params. &\multicolumn{2}{c}{Rain100L}\\
        \midrule
        \rowcolor{gray!10}DIDMDN&-&23.79&.773\\
        UMR & - & 32.39 & .921 \\
        \rowcolor{gray!10}SIRR & -& 32.37&.926 \\
        MSPFN & - & 33.50 & .948 \\
        \rowcolor{gray!10}LPNet  & - &  23.15 & .921 \\
        \midrule
        AirNet  & 9M & 34.90 & .977 \\
        \rowcolor{gray!10}PromptIR & 36M & {37.04} & 
        .979 \\
        \midrule 
        \rowcolor{green!2}\ourmethod (\textit{ours}) & \textcolor{tabred}{\textbf{6M}} & {37.47} & {.980}\\
        \rowcolor{green!2}\ourmethod(\textit{Ours}) & 10M & \textcolor{tabred}{\textbf{38.01}} & \textcolor{tabred}{\textbf{.982}} \\
        \bottomrule
        \end{tabularx}
    \end{subtable}%
    \hfill
    \hspace{2mm}
    \begin{subtable}[l]{0.55\textwidth}
        \subcaption{\textit{Denoising} on BSD68}
        \vspace{-1mm}
        \begin{tabularx}{\textwidth}{X*{7}{c}}
        \toprule
        Method & Params. & \multicolumn{2}{c}{$\sigma$=15} & \multicolumn{2}{c}{$\sigma$=25} & \multicolumn{2}{c}{$\sigma$=50}\\
        \midrule
        \rowcolor{gray!10}
        DnCNN&- & 33.89 & .930 & 31.23 & .883 & 27.92 & .789 \\
        IRCNN   & - & 33.87 & .929 & 31.18 & .882 & 27.88 & .790  \\
        \rowcolor{gray!10}
        FFDNet & - & 33.87 & .929 & 31.21 & .882 & 27.96 & .789 \\
        \midrule 
        BRDNet & - & 34.10 & .929 & 31.43 & .885 & 28.16 & .794 \\ 
        \rowcolor{gray!10}
        AirNet & 9M & 34.14 & .936 & 31.48 & .893 & 28.23 & .806  \\
        PromptIR & 36M & {34.34} & {.938} & 31.71 & .897 & 28.49 & .813 \\
        \rowcolor{gray!10}
        PromptIR (Reproduce) & 36M & {34.15} & {.934} & 31.50 & .894 & 28.33 & .807 \\
        \midrule
        \rowcolor{green!2}\ourmethod (\textit{ours})  & \textcolor{tabred}{\textbf{6M}} & 
        {34.23} & {.936} & {31.60} & {.896} & {28.36} & {.808} \\
        \rowcolor{green!2}\ourmethod(\textit{Ours}) & 10M & \textcolor{tabred}{\textbf{34.25}} & \textcolor{tabred}{\textbf{.937}} & \textcolor{tabred}{\textbf{31.65}} & \textcolor{tabred}{\textbf{.898}} & \textcolor{tabred}{\textbf{28.38}} & \textcolor{tabred}{\textbf{.810}} \\
        \bottomrule
        \end{tabularx}
    \end{subtable}
    }
    \vspace{0mm}
\end{table}

\noindent\textbf{Single-Degradation.}
In Tab.~\ref{tab:exp:single}, we compare our method against state-of-the-art approaches on single degradation tasks. For dehazing on SOTS dataset, we compare with DehazeNet~\cite{cai2016dehazenet}, MSCNN~\cite{ren2016single}, AODNet~\cite{li2017aod}, EPDN~\cite{qu2019enhanced}, FDGAN~\cite{dong2020fdgan}, and all-in-one methods AirNet~\cite{li2022all} and PromptIR~\cite{potlapalli2023promptir}. Our 6M parameter model achieves competitive performance (31.46 dB PSNR, 0.977 SSIM), while our 10M model establishes new state-of-the-art results (31.53 dB PSNR, 0.980 SSIM), outperforming the much larger PromptIR (36M parameters). For deraining on Rain100L, we evaluate against DIDMDN~\cite{zhang2018density}, UMR~\cite{yasarla2019uncertainty}, SIRR~\cite{wei2019semi}, MSPFN~\cite{jiang2020multi}, LPNet~\cite{gao2019dynamic}, AirNet~\cite{li2022all}, and PromptIR~\cite{potlapalli2023promptir}. Our method significantly outperforms all baselines, with our 10M model achieving 38.01 dB PSNR and 0.982 SSIM. For denoising on BSD68, we compare with classical methods DnCNN~\cite{zhang2017beyond}, IRCNN~\cite{zhang2017learning}, FFDNet~\cite{zhang2018ffdnet}, BRDNet~\cite{tian2000brdnet}, and recent all-in-one approaches AirNet~\cite{li2022all} and PromptIR~\cite{potlapalli2023promptir}. Our method consistently outperforms all competitors across different noise levels ($\sigma$=15, 25, 50), demonstrating superior performance with significantly fewer parameters than existing all-in-one methods.

\begin{algorithm}[t]
\caption{DynamicDepthwiseConv}
\label{alg:dynamic_depthwise_conv}
\begin{algorithmic}[1]
\Require $\alpha \in \mathbb{R}^{B \times C \times H \times W}$ \Comment{Input feature map}
\Ensure $\alpha' \in \mathbb{R}^{B \times C \times H \times W}$ \Comment{Output after dynamic depthwise conv}

\Statex \textbf{[Step 1] Generate Dynamic Kernel}
\State $K \gets \texttt{AdaptiveAvgPool2D}(\alpha)$ \Comment{Global context pooling}
\State $K \gets \texttt{Conv2D}(K,\ 1 \times 1,\ \text{out\_ch}=C)$ \Comment{Linear projection}
\State $K \gets \texttt{GELU}(K)$ \Comment{Non-linear activation}
\State $K \gets \texttt{Conv2D}(K,\ 1 \times 1,\ \text{out\_ch}=C \cdot k^2)$ \Comment{Generate kernel weights}
\State $K \gets \texttt{Reshape}(K,\ [B \cdot C,\ 1,\ k,\ k])$ \Comment{Form depthwise filters}

\Statex \textbf{[Step 2] Apply Depthwise Convolution}
\State $\alpha_{\text{flat}} \gets \texttt{Reshape}(\alpha,\ [1,\ B \cdot C,\ H,\ W])$ \Comment{Prepare for grouped conv}
\State $\alpha'_{\text{flat}} \gets \texttt{Conv2D}(\alpha_{\text{flat}},\ K,\ \text{groups}=B \cdot C,\ \text{padding}=k\div2)$ \Comment{Apply dynamic depthwise conv}
\State $\alpha' \gets \texttt{Reshape}(\alpha'_{\text{flat}},\ [B,\ C,\ H,\ W])$ \Comment{Reshape back to original shape}

\State \Return $\alpha'$
\end{algorithmic}
\end{algorithm}

\begin{figure}[!t]
    \centering
    \includegraphics[width=1.0\linewidth]{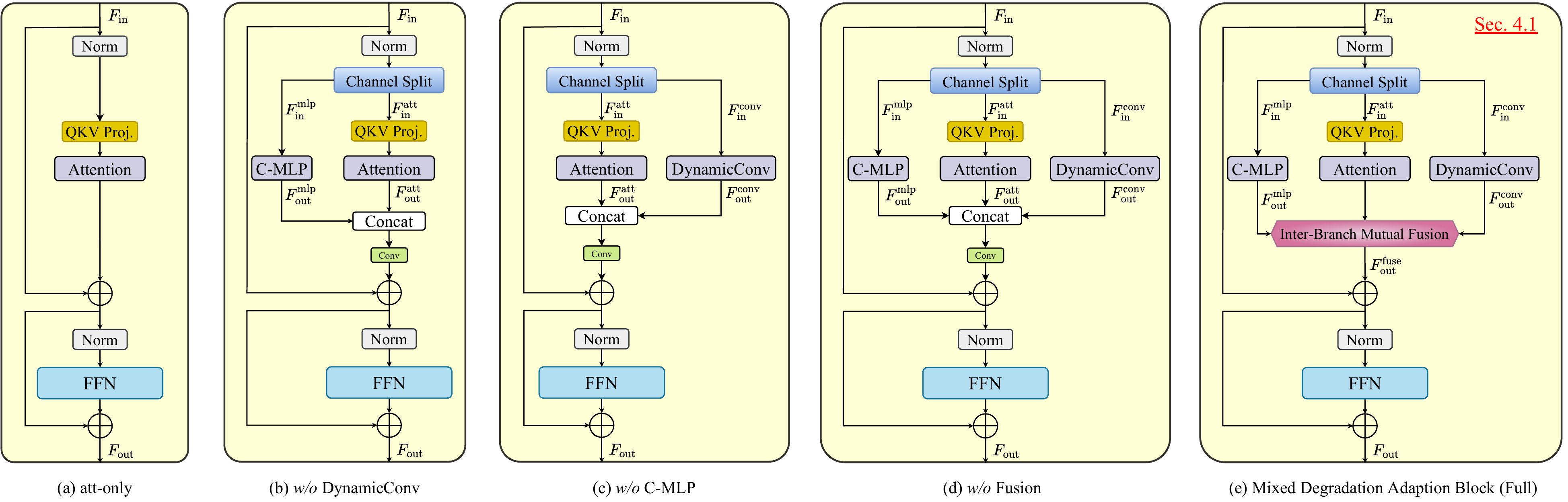}
    \vspace{-3mm}
    \caption{ The illustration of different designs of the proposed MDAB.}
    \label{fig:supp:ab_mdab}
    \vspace{-2mm}
\end{figure}
\subsection{Details of the Design for the proposed Mixed Backbone.}
To investigate the effectiveness of combining MLP, convolution, and attention mechanisms, we conducted an extensive design-level ablation study. The quantitative results are presented in Tab. 7 of the main manuscript. Here, we provide detailed visual illustrations of each design in Fig.~\ref{fig:supp:ab_mdab}.

\textbf{C-MLP.} To strengthen channel-wise representation, we introduce a Channel-wise MLP module, denoted as $\operatorname{C\text{-}MLP()}$. Given the input feature map $F_{\text{in}}^{\text{mlp}} \in \mathbb{R}^{B \times C \times H \times W}$, we first flatten the spatial dimensions to obtain a sequence $F_{\text{in}}^{\text{mlp}} \in \mathbb{R}^{B \times C \times L}$, where $L = H \times W$. The C-MLP is implemented using two 1D convolutional layers with a GELU activation in between. The GELU function introduces non-linearity, enabling the model to learn more complex and expressive channel-wise transformations. After processing, the output is reshaped back to the original spatial format, yielding $F_{\text{out}}^{\text{mlp}} \in \mathbb{R}^{B \times C \times H \times W}$.

\textbf{Dynamic Depthwise Convolution.} The $\operatorname{DynamicDepthwiseConv()}$ module is designed to capture content-adaptive local structures and is employed in Alg.1 of our main manuscript. As detailed in Alg.~\ref{alg:dynamic_depthwise_conv}, the input feature $\alpha \in \mathbb{R}^{B \times C \times H \times W}$ is first passed through a global average pooling and two $1 \times 1$ convolutions to generate a dynamic depthwise kernel for each channel and sample. The input is reshaped and convolved with the generated kernels using grouped convolution, enabling sample-specific spatial filtering. The resulting output $\alpha'$ maintains the original resolution while embedding adaptive local information.

\begin{algorithm}[t]
\caption{SPD Contrastive Learning Optimization Pseudocode}
\label{alg:code}
\vspace{-1mm}
\begin{lstlisting}[language=python]
# (*@$f_{en}$@*): encoder
# (*@$f_{de}$@*): decoder
# (*@$patch\_embedding$@*): shallow convolutional patch embedding
# (*@$refinement\_conv$@*): the refinement block and the final convolution
# (*@$spd$@*): compute SPD feature
for (*@${x}$@*) in loader:  # load a minibatch x with n samples

    (*@$ F_{\text{shallow}} $@*) = (*@$patch\_embedding$@*)((*@${x}$@*))  # Convolutional Patch Embedding
    (*@$ F_{\text{latent}} $@*) = (*@$f_{en}$@*)((*@$F_{\text{shallow}}$@*))

    (*@$\mathbf{C}_{s}$@*), (*@$\mathbf{C}_{l}$@*)= (*@$spd$@*)((*@$F_{\text{shallow}}$@*)), (*@$spd$@*)((*@$F_{\text{latent}}$@*)) # Compurte SPD (Symmetric Positive Definite) manifold features
    (*@$\mathbf{z}_{s}$@*), (*@$\mathbf{z}_{l}$@*) = (*@$proj\_norm$@*)((*@$\mathbf{C}_{s}$@*)), (*@$proj\_norm$@*)((*@$\mathbf{C}_{l}$@*)) # Projection and normalize

    (*@$ F_{\text{recon}} $@*) = (*@$f_{de}$@*)((*@$F_{\text{latent}}$@*))
    (*@$\hat{x}$@*) = (*@$refinement\_conv$@*)((*@$ F_{\text{recon}} $@*))
    
    L = (*@$\mathcal{L}_{\text{1}}$@*)((*@${x}$@*), (*@$\hat{x}$@*)) + (*@$\lambda_{fre} \times $@*)(*@$\mathcal{L}_{\text{Fourier}}$@*) ((*@${x}$@*), (*@$\hat{x}$@*)) + (*@$\lambda_{ctrs} \times $@*)(*@$\mathcal{L}_{\text{SPD}}$@*)((*@$\mathbf{z}_{s}$@*), (*@$\mathbf{z}_{l}$@*)) # total loss
    
    L.backward()  # back-propagate
    update((*@$f_{en}$@*), (*@$f_{de}$@*), (*@$patch\_embedding$@*), (*@$refinement\_conv$@*))  # SGD update

    
def (*@$\mathcal{L}_{\text{Fourier}}$@*)(a, b):  #  Real-valued Fourier loss

    Please refer to Eq.B of our Appendix.

    return loss

def (*@$\mathcal{L}_{\text{SPD}}$@*)(a, b):  #  SPD Loss
    
    Please refer to Eq.5 of our main manuscript.
    
    return loss
\end{lstlisting}
\vspace{-1mm}
\end{algorithm}
\subsection{Details of the Proposed SPD Contrastive Learning.}
\label{subsec:supp:spd_loss}
As shown in Alg.~\ref{alg:code}, our SPD-based contrastive learning aims to align shallow and latent representations by operating in the space of symmetric positive definite (SPD) matrices. Specifically, given the shallow features extracted from the convolutional patch embedding and the latent features produced by the encoder, we compute their second-order channel-wise statistics to obtain SPD representations. These matrices are then vectorized and projected through learnable MLP layers, followed by $\ell_2$ normalization to form contrastive embeddings. An InfoNCE-style loss is applied between the shallow and latent embeddings to encourage structural alignment across depth. This contrastive term complements the pixel-level and frequency-based objectives, promoting more discriminative and consistent feature learning without introducing any additional cost during inference.
Importantly, by leveraging the geometry of second-order feature statistics, our approach implicitly regularizes the representation space, encouraging intra-instance compactness and inter-degradation separability. This geometrically grounded formulation bridges low-level signal priors with high-level contrastive learning, offering a principled and scalable solution to all-in-one image restoration.

\subsection{Ablation Regarding the Optimization Objectives}
Tab.~\ref{tab:supp:ab_loss} shows that replacing SPD-based contrastive learning with a standard Euclidean-space contrastive loss (\textit{w/o SPD}) results in a clear performance drop, 
\begin{wraptable}{r}{0.49\textwidth}
    \centering
    \scriptsize
    \setlength\tabcolsep{9pt}
    \setlength{\extrarowheight}{0.05pt}
    \vspace{0mm}
    \caption{\textit{Ablation Study} of \ourmethod-T on 3 Degradation Setting.}
    \label{tab:supp:ab_loss}
    \vspace{-3mm}
    \scalebox{0.89}{
    \begin{tabular}{lccc}
    \toprule
    \multirow{2}{*}{Ablaton} & \multirow{2}{*}{Parms.} &  \multicolumn{2}{c}{Results} \\
    \cmidrule(lr){3-4} && PSNR (dB, $\uparrow$) & SSIM($\downarrow$) \\
    \midrule
    \rowcolor{gray!10}
    \textit{w/o} CL \& SPD & 5.80M & 32.63 (\sotab{-0.14}) & .916\\
    \textit{w/o} SPD & 6.10M & 32.53 (\sotab{-0.24}) & .914 \\
    \midrule
    \textit{w/o} Fourier Loss & 5.80M & 32.70 (\sotab{-0.07}) & .917\\
    \midrule
    \rowcolor{green!2}\ourmethod-T \textit{(Full) }& 6.21M & \sotaa{32.77} & \sotaa{.919} \\
    \bottomrule
    \end{tabular}
    }
    \vspace{-2mm}
\end{wraptable}
demonstrating the advantage of modeling second-order channel correlations on the SPD manifold rather than relying solely on first-order vector similarities. 
When the entire contrastive module is removed (\textit{w/o CL \& SPD}), performance degrades even further, indicating that aligning shallow and deep features is essential for effective representation learning. 
Moreover, removing the Fourier loss (\textit{w/o Fourier Loss}) slightly reduces performance, suggesting that frequency-domain supervision provides additional benefits. Overall, the full model achieves the best results, confirming the effectiveness of jointly optimizing spatial, frequency, and SPD-manifold-based structural consistency. Note that throughout all the experiments, we set \(\lambda_{ctrs}\) = 0.05 and \(\lambda_{ctrs}\)=0.1.

\subsection{Shallow-Latent Feature Similarity}
Besides the channel-wise similarity comparison provided in our main manuscript for denoising. We also find consistent findings in other degradation, \ie, raining and hazing. The corresponding channel-wise similarity across scales is provided in Fig.~\ref{fig:supp:sim_vis_3deg}. 
\begin{figure}[!t]
    \centering
    \includegraphics[width=1.0\linewidth]{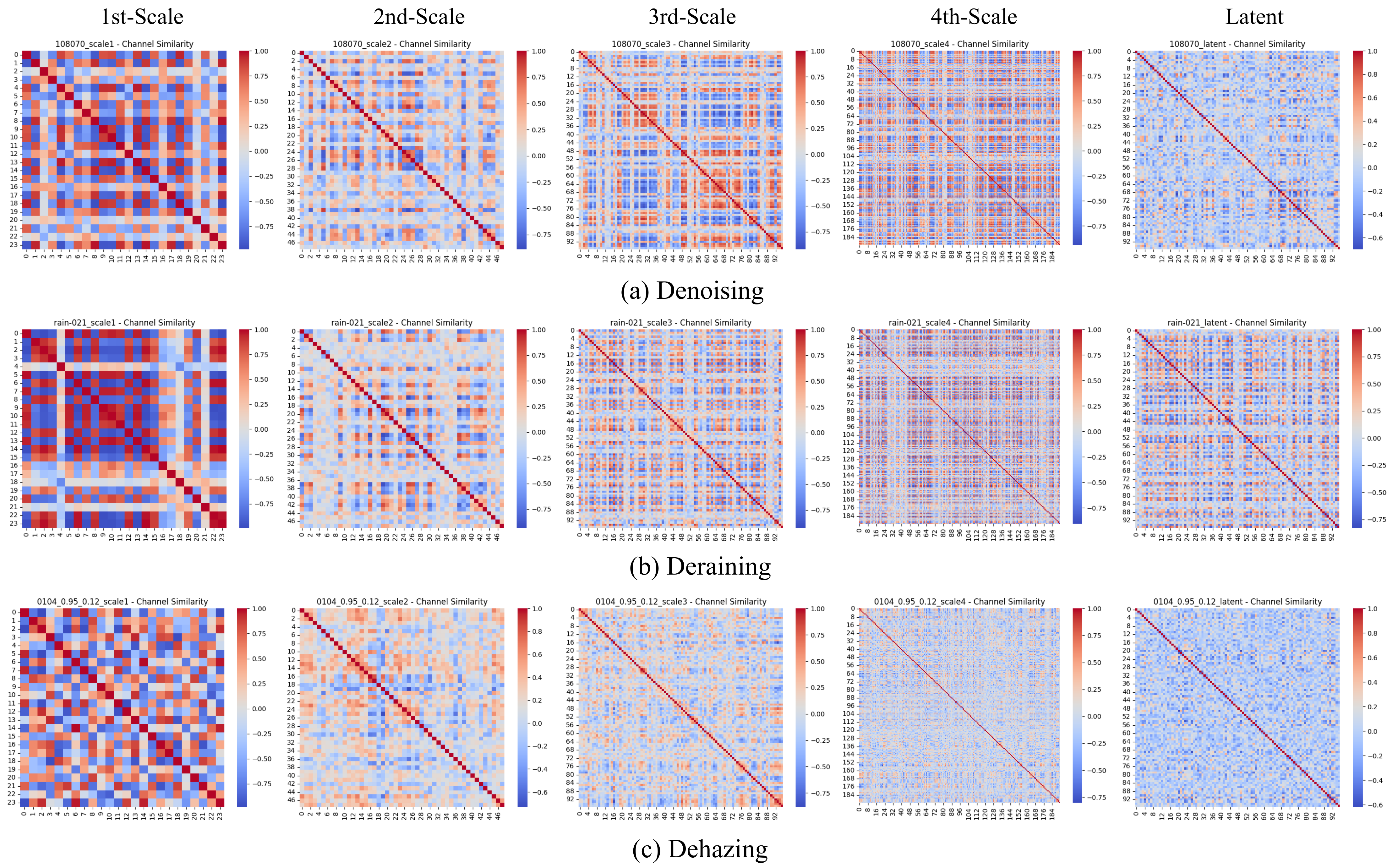}
    \vspace{-3mm}
    \caption{The cross-sclae channel-wise similarity matrix visualization for Denoising, Deraining, and Dehazing.}
    \label{fig:supp:sim_vis_3deg}
    \vspace{-2mm}
\end{figure}
These observations reveal several important trends:
\textit{(i)} Despite the diversity of degradation types, a consistent pattern emerges across scales. Specifically, from the first to the fourth scale, the overall channel-wise similarity indicates substantial redundancy among feature channels. After channel reduction, the latent features become more decorrelated, which validates the rationale for applying contrastive learning between the latent and shallow (\ie, first-scale) features.
\textit{(ii)} Different degradation types exhibit varying degrees of channel redundancy. As illustrated in Fig.~\ref{fig:supp:sim_vis_3deg}, hazy images tend to produce more inherently independent features, whereas rain-degraded inputs show strong channel-wise redundancy even in the latent space. This suggests that degradations like haze may benefit from larger embedding dimensions to capture more expressive representations, while simpler degradations (\eg, rain) can achieve effective restoration with smaller embedding sizes due to their inherently redundant structure. 

These insights open up new directions for adaptive and degradation-aware model design in future research.
Notably, this trend is not limited to the three representative samples shown; we observe similar patterns consistently across the dataset in a statistical sense. We plan to conduct a more comprehensive and quantitative investigation of this phenomenon in future work.

\subsection{More Generlization Evaluation}
\label{suppsec:unseen}
\begin{table}[!t]
    \centering
    \scriptsize
    \caption{Zero-shot evaluation on real-world under-display camera datasets TOLED and POLED~\citep{zhou2021image}.}
    \label{tab:exp:udc_zero_shot}
    \vspace{-2mm}
    \setlength{\tabcolsep}{4pt}
    \renewcommand{\arraystretch}{1.1}
    \begin{tabular}{lcc}
    \toprule
    \textbf{Method} & \textbf{TOLED (PSNR / SSIM / LPIPS)} & \textbf{POLED (PSNR / SSIM / LPIPS)} \\
    \midrule
    AirNet~\citep{li2022all} & 14.58 / 0.609 / 0.445 & 7.53 / 0.350 / 0.820 \\
    PromptIR~\citep{potlapalli2023promptir} & 16.70 / 0.688 / 0.422 & 13.16 / 0.583 / 0.619 \\
    DiffUIR~\citep{zheng2024selective} & \textbf{29.55} / \textbf{0.887} / \textbf{0.281} & 15.62 / 0.424 / 0.505 \\
    \rowcolor{gray!10}
    \textbf{MIRAGE-S (Ours)} & 28.01 / {0.881} / {0.293} & \textbf{16.93} / \textbf{0.604} / \textbf{0.500} \\
    \bottomrule
    \end{tabular}
\end{table}

To further assess generalization beyond synthetic settings, we evaluate MIRAGE-S on the real-world TOLED and POLED under-display camera datasets~\citep{zhou2021image}. As shown in Tab.~\ref{tab:exp:udc_zero_shot}, MIRAGE-S achieves strong performance across both benchmarks. On POLED, which contains more severe signal attenuation and non-linear spatial artifacts, MIRAGE-S clearly surpasses prior methods across all three metrics, indicating robust transfer to challenging real-world degradations. On TOLED, MIRAGE-S remains competitive and delivers results close to diffusion-based DiffUIR despite its significantly lower complexity. These findings suggest that the proposed mixed-backbone architecture and SPD-based alignment maintain good stability under real sensor degradations and generalize reliably across distinct UDC hardware conditions.

\section{Additional Visual Results.}
\label{suppsec:vis}
\subsection{3 Degradation}
\begin{figure}[!t]
    \centering
    \includegraphics[width=1.0\linewidth]{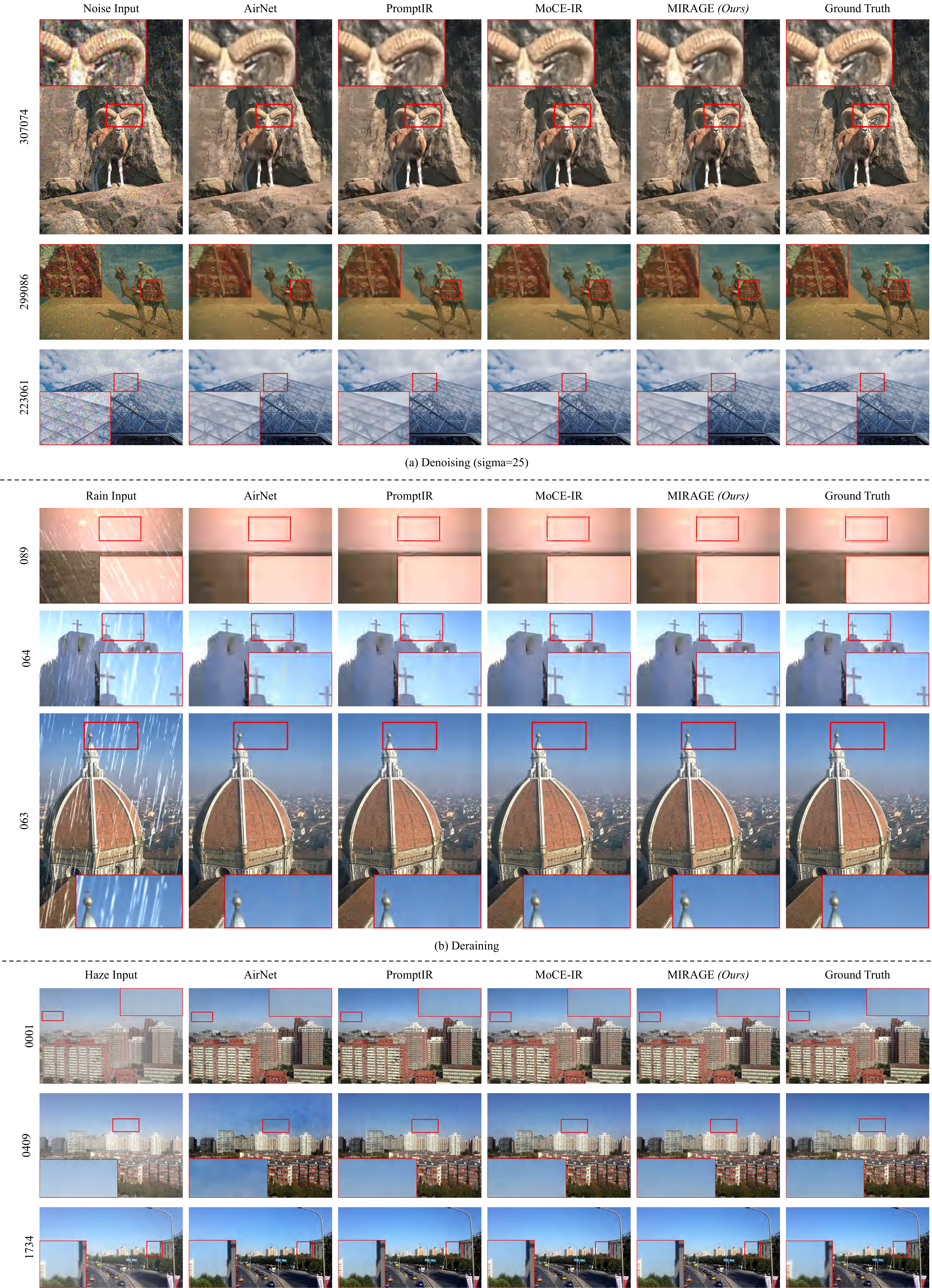}
    \vspace{-3mm}
    \caption{Visual comparison of \ourmethod with state-of-the-art methods considering three degradations. Zoom in for a better view.}
    \label{fig:supp:vis_3deg}
    \vspace{-1mm}
\end{figure}
Fig.~\ref{fig:supp:vis_3deg} presents qualitative comparisons on representative cases of denoising, deraining, and dehazing, benchmarked against recent state-of-the-art methods. The proposed \ourmethod consistently yields more visually faithful restorations, characterized by enhanced structural integrity, finer texture details, and reduced artifacts. These results underscore the effectiveness of our unified framework in handling diverse degradation types while preserving high-frequency information and geometric consistency.

\subsection{5 Degradation}
\begin{figure}[!t]
    \centering
    \includegraphics[width=1.0\linewidth]{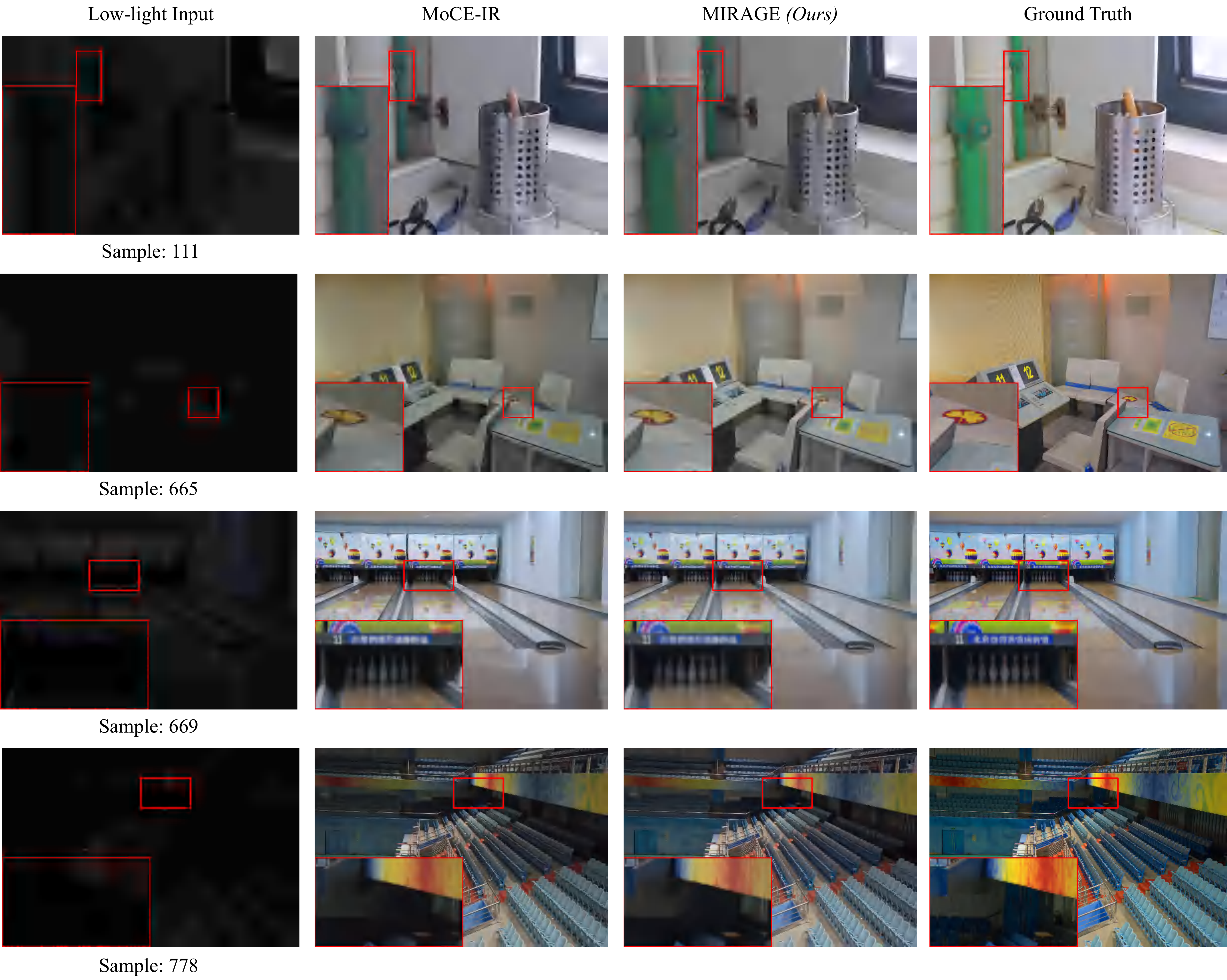}
    \vspace{-4mm}
    \caption{Visual comparison of \ourmethod with state-of-the-art methods considering low-light degradation. Zoom in for a better view.}
    \label{fig:supp:vis_lowlight}
    \vspace{-2mm}
\end{figure}
For the 5-degradation setting, we provide visual comparisons for the low-light enhancement task in Fig.~\ref{fig:supp:vis_lowlight}. As illustrated, the proposed \ourmethod produces noticeably cleaner outputs with improved luminance restoration and better color consistency compared to MoCE-IR\cite{zamfir2025moce}, demonstrating its robustness under challenging illumination conditions.

\subsection{Composited Degradation}
\begin{figure}[!t]
    \centering
    \includegraphics[width=1.0\linewidth]{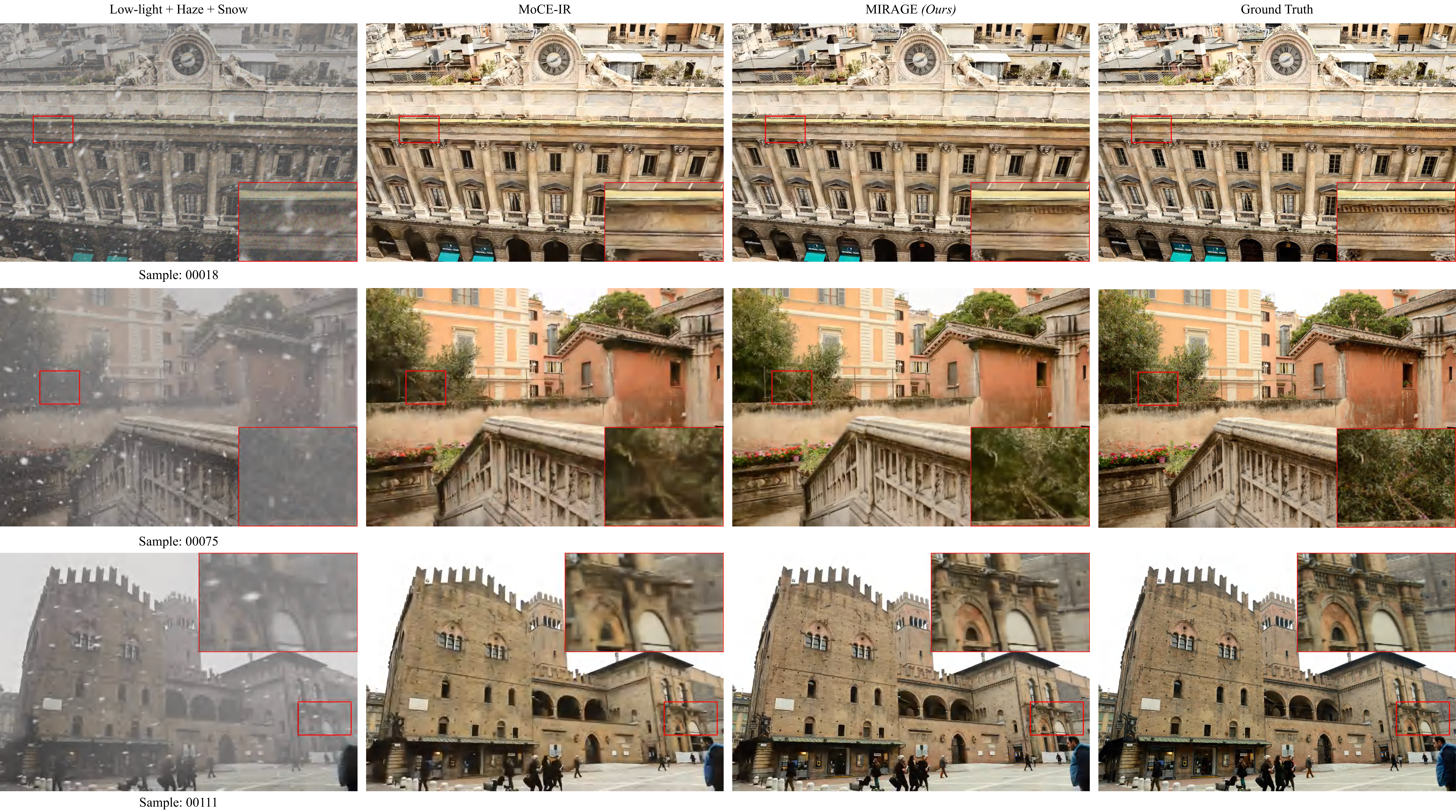}
    \vspace{-4mm}
    \caption{Visual comparison of \ourmethod with state-of-the-art methods considering composited degradation (Low-light + Haze + Snow). Zoom in for a better view.}
    \label{fig:supp:vis_lhs}
    \vspace{-1mm}
\end{figure}

\begin{figure}[!t]
    \centering
    \includegraphics[width=1.0\linewidth]{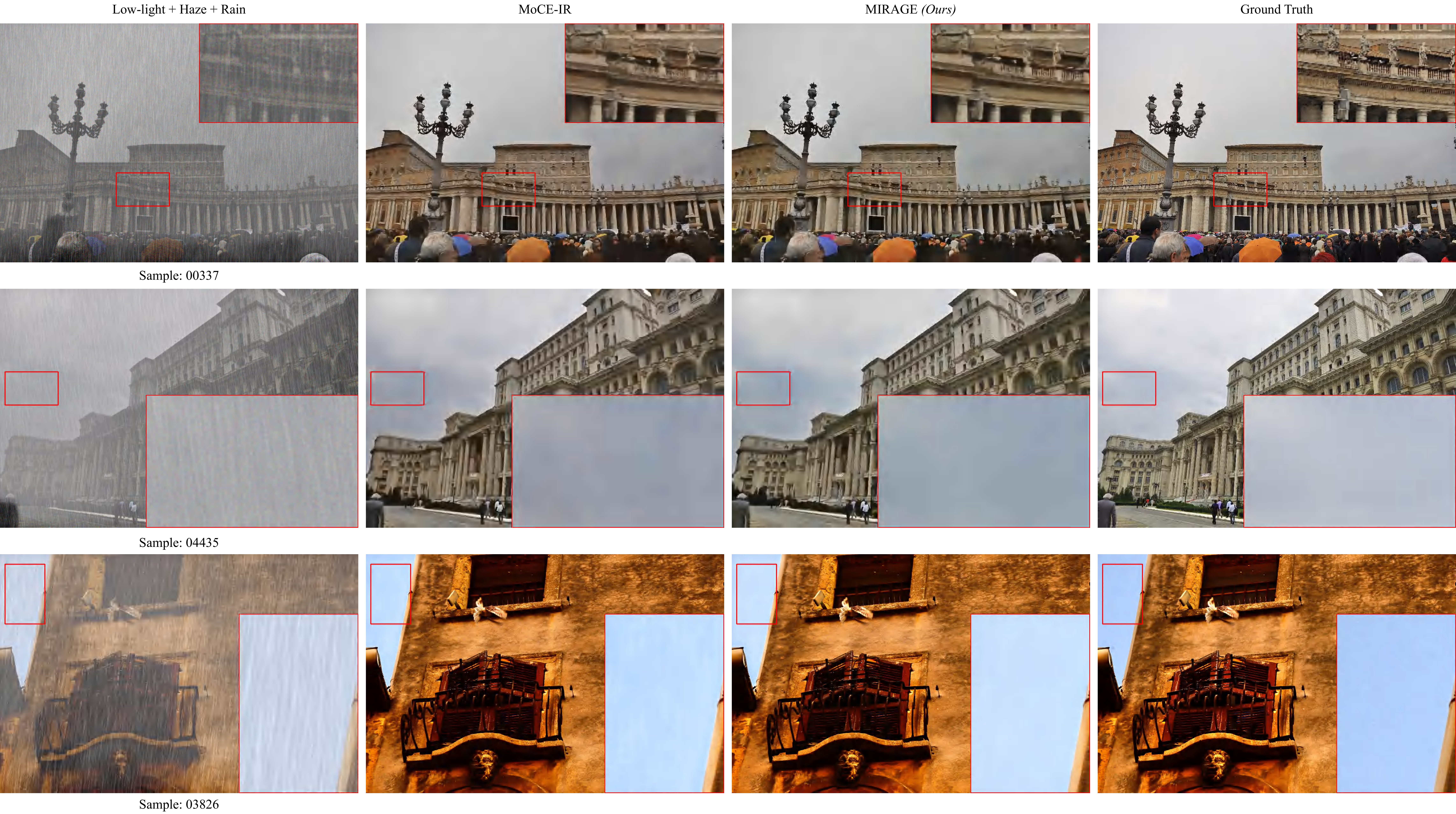}
    \vspace{-4mm}
    \caption{Visual comparison of \ourmethod with state-of-the-art methods considering composited degradation (Low-light + Haze + Rain). Zoom in for a better view.}
    \label{fig:supp:vis_lhr}
    \vspace{-2mm}
\end{figure}

Fig.~\ref{fig:supp:vis_lhs} and Fig.~\ref{fig:supp:vis_lhr} present visual comparisons under more challenging composite degradations, namely \textit{low-light + haze + snow} and \textit{low-light + haze + rain}, respectively. As observed, our method reconstructs significantly more scene details and preserves structural consistency, whereas MoCE-IR~\cite{zamfir2025moce} tends to produce noticeable artifacts and over-smoothed regions under these complex conditions.

\subsection{Zero-Shot Underwater Image Enhancement}
\begin{figure}[!t]
    \centering
    \includegraphics[width=1.0\linewidth]{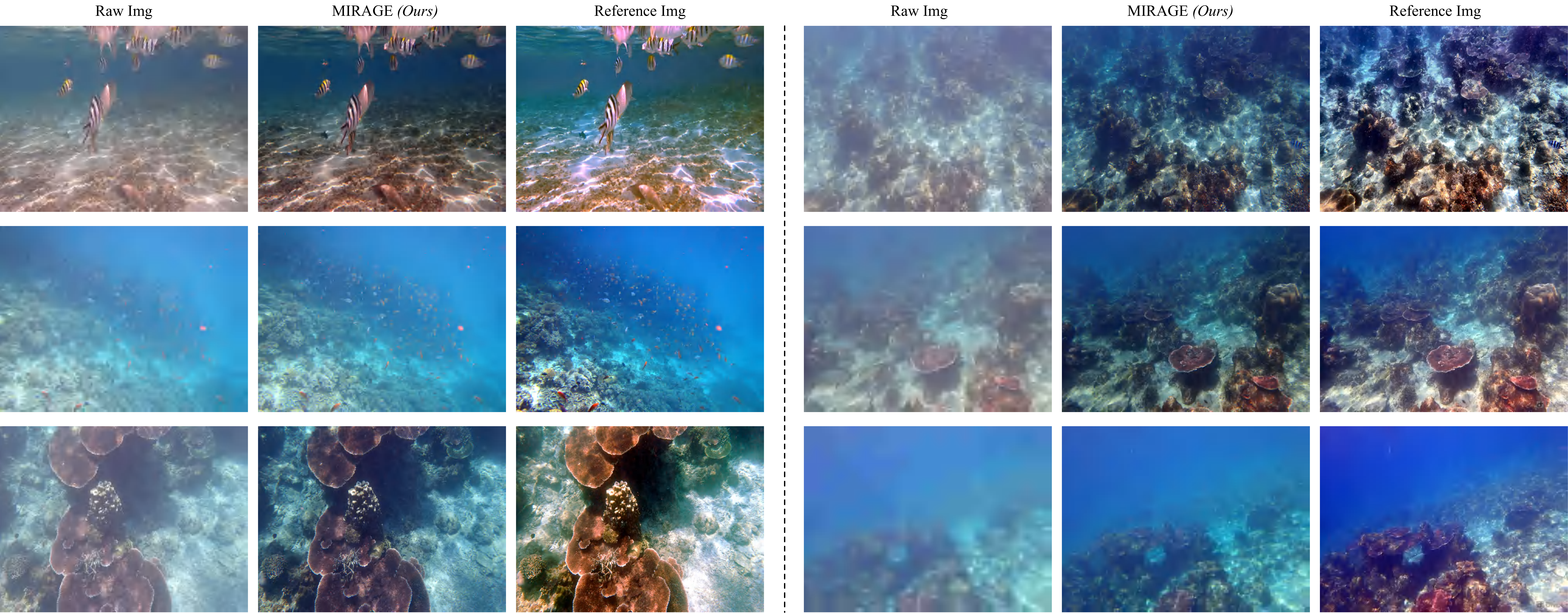}
    \vspace{-4mm}
    \caption{Visual results of \ourmethod for Underwater Image Enhancement. Zoom in for a better view.}
    \label{fig:supp:vis_udie}
    \vspace{0mm}
\end{figure}
Fig.~\ref{fig:supp:vis_udie} demonstrates that even when directly applied to unseen underwater images, our method is able to effectively enhance visibility and contrast, producing results that are noticeably clearer than the raw input and visually closer to the reference images. This qualitative evidence further validates the strong generalization ability of the proposed framework to unseen domains.

\section{Limitations and Future Work}
\label{suppsec:limit}
While the proposed \ourmethod achieves new state-of-the-art performance on most all-in-one image restoration benchmarks, we observe that its deblurring performance still lags slightly behind MoCE-IR~\cite{zamfir2025moce}. We attribute this to the relatively compact model size of our current design, which favors efficiency over aggressive capacity.
To address this, future work will explore scaling up the model size to be on par with larger architectures such as PromptIR~\cite{potlapalli2023promptir}, MoCE-IR~\cite{zamfir2025moce}, and AdaIR~\cite{cui2025adair}, aiming to further boost performance while maintaining the architectural elegance and efficiency of our design.
Moreover, our current SPD-based contrastive learning leverages a conventional InfoNCE loss in Euclidean space after projecting SPD features. While effective, it does not fully exploit the intrinsic geometry of the SPD manifold. As part of future efforts, we plan to investigate geodesic-based contrastive formulations and Riemannian-aware optimization strategies, which may offer a more principled and theoretically grounded way to align structured representations across semantic scales.
Additionally, different degradations may favor different proportions of convolution, attention, and MLP capacity. Learning such ratios dynamically is an interesting direction and could further adapt MIRAGE to degradation-specific characteristics. We view this as a promising avenue for future research.

\section{Broader Impact}
\label{suppsec:impact}
Image restoration (IR) is a fundamental task with applications in photography, remote sensing, surveillance, autonomous driving, medical imaging, and scientific visualization. By proposing a unified and efficient framework capable of handling diverse degradation types with minimal computational cost, our work may benefit scenarios where image quality is compromised by environmental or hardware constraints. 
The lightweight design of MIRAGE further enables deployment on resource-limited devices such as mobile phones, drones, or embedded cameras, which can support use cases in low-resource settings or critical domains like emergency response and environmental monitoring. From a research perspective, our modular design and SPD-based contrastive formulation may encourage further exploration of geometrically-aware representation learning in restoration and related areas.

\section{Use of Large Language Models (LLMs)}
\label{suppsec:llm}
We used OpenAI's GPT-based Large Language Models (LLMs)~\citep{openai2023gpt4,ChatGPT} to polish the writing and improve the readability of the paper. The models were not used for developing the methodology, running experiments, or analyzing results. All scientific contributions remain entirely the work of the authors.

\end{document}